%% file: main.tex
\documentclass[sigconf]{acmart}

\AtBeginDocument{%
  }

\setcopyright{acmlicensed}
\copyrightyear{2026}
\acmYear{2026}
\acmDOI{XXXXXXX.XXXXXXX}

\usepackage{enumitem}
\usepackage{amsmath}
\usepackage[normalem]{ulem}
\usepackage{xspace}
\usepackage{array}
\usepackage{makecell}
\usepackage{booktabs}
\usepackage{longtable}
\usepackage{tabularx}
\usepackage{ragged2e}
\usepackage{caption}
\usepackage{multirow}
\usepackage{algorithm}
\usepackage{algpseudocode}
\usepackage{listings}
\usepackage[most]{tcolorbox}
\usepackage{graphicx}
\usepackage{hyperref}
\usepackage{url}
\usepackage{placeins}

\newcolumntype{L}[1]{>{\raggedright\arraybackslash}m{#1}}
\newcolumntype{C}[1]{>{\centering\arraybackslash}m{#1}}

\input{cmd}

\begin{document}

\input{title}

\input{authors}
\input{0_abstract}

\maketitle

\input{1_intro_new}
\input{2_related_work}
\input{3_sys_overview}
\input{4_plover}

\input{5_evaluation}

\input{6_discussion}

\input{7_conclusion}

\begin{acks}
This work was supported in part by the U.S. National Science Foundation under Grant No. IIS-2427770 and by a Bosch Research Award.
\end{acks}

\bibliographystyle{ACM-Reference-Format}
\bibliography{main}

\input{8_appendix}

\end{document}

%% file: cmd.tex
\definecolor{dypink}{HTML}{ec008c}
\definecolor{dypurple}{HTML}{8654d1}

\newif\ifnotes
\notestrue 

\usepackage{amssymb}

\usepackage[most]{tcolorbox}
\definecolor{cback}{RGB}{237,239,241}
\definecolor{cframe}{RGB}{185,196,202}
\definecolor{cgrey}{RGB}{102,102,102}
\definecolor{cL}{RGB}{186,230,253}
\definecolor{cQ}{RGB}{229,217,242}
\definecolor{cG}{RGB}{216,247,147}

\newtcbox{\inlineGbox}[1][]{enhanced,
 box align=base,
 nobeforeafter,
 colback=cG,
 colframe=cframe,
 size=small,
 fontupper=\footnotesize,
 left=1pt,
 right=1pt,
 boxsep=0.5pt,
 #1}

 \newtcbox{\inlineLbox}[1][]{enhanced,
 box align=base,
 nobeforeafter,
 colback=cL,
 colframe=cframe,
 size=small,
 fontupper=\footnotesize,
 left=1pt,
 right=1pt,
 boxsep=0.5pt,
 #1}

 \newtcbox{\inlineQbox}[1][]{enhanced,
 box align=base,
 nobeforeafter,
 colback=cQ,
 colframe=cframe,
 size=small,
 fontupper=\footnotesize,
 left=1pt,
 right=1pt,
 boxsep=0.5pt,
 #1}

\usepackage{xcolor}
\definecolor{ploverblack}{RGB}{0,0,0}
\definecolor{ploverpurple}{RGB}{75, 0, 130}

\newcommand{\maincontrib}[1]{\textcolor{ploverblack}{\texttt{#1}}}

\newcommand{\intervention}[1]{\textcolor{ploverpurple}{\textbf{\textit{#1}}}}

\newcommand{\feature}[1]{\textbf{#1}}

\newcommand{\mech}[1]{\textit{#1}}

\usepackage{tikz}
\definecolor{circlecolorpink}{RGB}{171, 87, 112}

\newcommand{\circlep}[1]{%
\tikz[baseline=(char.base)]{
\node[shape=circle,draw=circlecolorpink,inner sep=1pt, fill=circlecolorpink, text=white] (char) {#1};}%
}
\definecolor{circlecolorgreen}{RGB}{86, 94, 67}

\newcommand{\circleg}[1]{%
\tikz[baseline=(char.base)]{
\node[shape=circle,draw=circlecolorgreen,inner sep=1pt, fill=circlecolorgreen, text=white] (char) {#1};}%
}

\usepackage{pifont}
\usepackage{wasysym}
\newcommand{\successmark}{
\tikz[baseline=-0.5ex]\node[
draw=none,
fill=green!70!black,
circle,
inner sep=1.5pt
]{\textcolor{white}{\small\ding{51}}};
}

\newcommand{\failmark}{
\tikz[baseline=-0.5ex]\node[
draw=none,
fill=red!75!black,
circle,
inner sep=1.5pt
]{\textcolor{white}{\small\ding{55}}};
}
\definecolor{purpletag}{RGB}{125, 84, 154}
\newtcbox{\purpletag}{
enhanced,
colback=purpletag,
coltext=white,
boxrule=0pt,
arc=2pt,
left=3pt,
right=3pt,
top=2pt,
bottom=2pt,
valign=center,
tcbox raise base,
center,
width=2.4cm,   
}

\usepackage[table]{xcolor}
\usepackage{array}
\usepackage{colortbl}

\definecolor{headerbg}{HTML}{c1b7d1}
\definecolor{stripebg}{HTML}{f4f3f7}
\definecolor{gridclr}{HTML}{cac9d3}

%% file: title.tex
\newcommand{\sys}{\textsc{Plover}}
\newcommand{\syss}{\textsc{Plover}\xspace}

\title{\sys: Steering GUI Agents through Plan-Centric Interaction}


%% file: authors.tex
\author{Madhumitha Venkatesan}
\affiliation{%
  \institution{University of California, Davis}
  \country{Davis, California, USA}
}
\email{mvenkat@ucdavis.edu}

\author{Shicheng Wen}
\affiliation{%
  \institution{University of California, Davis}
  \country{Davis, California, USA}
}
\email{wenshicheng@ucdavis.edu}

\author{Jiajing Guo}
\affiliation{%
  \institution{Bosch Research North America}
  \country{Sunnyvale, California, USA}
}
\email{jiajing.guo@ucdavis.edu}

\author{Jorge Piazentin Ono}
\affiliation{%
  \institution{Bosch Research North America}
  \country{Sunnyvale, California, USA}
}
\email{jorge.piazentinono@us.bosch.com}

\author{Liu Ren}
\affiliation{%
  \institution{Bosch Research North America}
  \country{Sunnyvale, California, USA}
}
\email{liu.ren@us.bosch.com}

\author{Dongyu Liu}
\affiliation{%
  \institution{University of California, Davis}
  \country{Davis, California, USA}
}
\email{dyuliu@ucdavis.edu}

\renewcommand{\shortauthors}{Venkatesan et al.}

%% file: 0_abstract.tex
\begin{abstract}
Graphical user interface (GUI) automation remains challenging in real-world environments, where dynamic layouts, unexpected dialogs, and evolving interface states can cause autonomous agents to drift from user intent. 
Recent vision-based multimodal agents improve flexibility by operating directly over screenshots and natural language instructions, but planning and adaptation often remain internal, limiting users' ability to inspect, supervise, or correct system behavior. 
We present \sys, a plan-centric vision-based GUI automation system that externalizes task plans and replanning as persistent, inspectable, and revisable artifacts. 
Through a planner--executor architecture, \syss supports explicit supervision of evolving execution, localized correction through editable plans, natural-language guidance, and screenshot-grounded interventions, while preserving prior progress during repair. 
A formative study with six participants informed the interaction design. We then evaluate \syss through benchmark failure-case repair and scenario-based workflow analyses. Our results show that many autonomous GUI-agent failures are structurally repairable when plans remain visible and interventions are localized, and that explicit replanning helps make GUI automation more transparent, controllable, and adaptable.
\end{abstract}

%% file: 1_intro_new.tex
\section{INTRODUCTION}
\label{intro}
Traditional rule-based scripts \cite{hellmann2011rule, qian2020roscript, granda2021towards} and commercial Robotic Process Automation (RPA) tools \cite{wewerka2020robotic, hofmann2020robotic} are brittle because they depend on fixed selectors, pixel anchors, or other heuristics tied to interface structure. Even minor UI changes can invalidate these bindings and require repeated manual repair \cite{pruucha2025llm}. Vision-based agents instead treat rendered pixels as the primary representation of interface state, acting through screenshots and observable mouse and keyboard actions rather than hidden structural metadata. This makes them particularly useful in heterogeneous or constrained environments, such as legacy tools, proprietary software, non-web applications, and interfaces whose structure is unstable or inaccessible. Recent multimodal large language models (MLLMs) with computer-use capabilities \cite{yao2022webshop, yang2024swe, team2026kimi, shen2025mind, zhao2025osagentssurvey} further extend this approach by integrating perception, planning, memory, and execution across diverse applications and workflows.

Despite these advances, most systems remain autonomy-first. Once execution begins, planning and replanning typically remain internal. Agents may silently reinterpret intent, regenerate steps, or pursue plausible but incorrect actions without maintaining shared situational awareness with the user \cite{shayegani2025just}. This creates a fundamental design tension: greater autonomy can improve flexibility, but it can also obscure system behavior and reduce user steerability when plans evolve under uncertainty. Accordingly, recent work has begun to explore collaboration and interaction as mechanisms for shared control in agentic systems \cite{chen2023gap, goyal2024designing, xu2025duetui, issak2025mosaaic, tang2025dark}. 
Yet important interface and interaction design gaps remain. First, task plans are often presented as one-shot outputs rather than persistent, editable artifacts. Second, current interfaces remain largely text-centric despite the visual and spatial nature of GUI tasks. Third, replanning is typically treated as an internal recovery mechanism rather than an explicit, inspectable interaction. Together, these limitations make moments of instability harder for users to detect, interpret, and repair. As a result, even advanced vision-based agents can still obscure the very breakdowns that call for human guidance. 

Moreover, high-level natural language correction alone is often insufficient because GUI failures are frequently situated, spatial, and state-dependent. A user may know that the agent clicked the wrong dropdown, selected the wrong cell, or skipped a modal dialog, but describing the exact target verbally can be ambiguous when multiple visually similar elements are present. Similarly, a corrective prompt often gives the agent new intent without specifying which part of the existing workflow should be preserved or revised. This can lead to broad regeneration, duplicated work, or loss of already completed progress. Plover therefore treats correction not only as a conversational instruction, but as a localized update to an explicit plan grounded in the current screen state. 

We do not suggest that every GUI task should require continuous user supervision. For short, routine, or low-risk tasks, fully autonomous execution or simple prompt re-issuance may be sufficient. The challenge lies in long-horizon, high-friction, or failure-prone workflows where errors may silently propagate, partial progress is valuable, and users already need some ability to verify outcomes. In such settings, the interaction challenge is not to replace delegation with manual control, but to make the necessary moments of supervision precise, localized, and recoverable. Hence, we argue that robust GUI automation should be treated not only as a modeling problem, but also as an interaction problem. When execution unfolds over multiple steps in dynamic interfaces, users need visibility into how the agent plans, adapts, and recovers over time. We therefore study GUI automation as a supervised, repairable process rather than a fully autonomous one \cite{li2017sugilite}.
To address the design gaps above, we introduce \sys, a plan-centric GUI automation system that externalizes task plans as persistent, inspectable, and revisable artifacts. Built on a vision-based GUI agent running in OS-level virtual machines, \syss surfaces plan changes explicitly and supports localized correction through editable plans, natural-language guidance, and screenshot-based interventions. This design provides a mechanism for users to supervise evolving execution, repair pending workflow segments, and preserve prior progress as tasks drift over time, rather than relying on repeated prompt re-issuance or hidden replanning.

We evaluate \syss through a formative study that informed the interaction design, followed by three complementary analyses of system behavior. In a benchmark failure-case repair analysis, we first ran 38 challenging tasks under autonomous execution, of which 26 resulted in non-successful outcomes. We then re-ran these 26 tasks in a collaborative setting and found that 23 improved, with 17 becoming complete successes and 6 becoming partial successes, requiring 2.04 interventions per task on average. We further conduct autonomy-only execution through scenario-based stability analysis using visual similarity and plan alignment metrics to examine replay stability and structural plan mismatch. Across both analyses, we identify recurring failure modes including spatial ambiguity in dense interfaces, execution drift in multi-step workflows, and misinterpreted intermediate states. These findings suggest that many GUI-agent failures are not simply terminal errors; they often become recoverable when plans are exposed, and intervention points are made explicit.

In summary, we contribute the following:
\begin{itemize}[leftmargin=10pt,topsep=0pt]
    \item A plan-centric interaction design for vision-based GUI automation that externalizes task plans as persistent, inspectable, and revisable artifacts, enabling explicit supervision of plan updates, localized repair, and visible replanning during execution. 
    \item \syss, a system that instantiates this design through editable plan artifacts, explicit \maincontrib{Intelligent Replanning}, and multimodal interventions that support targeted correction while preserving execution history and task continuity.
    \item An empirical characterization of expert-guided recoverability across benchmark GUI-agent failure cases and scenario-based workflows, showing where localized plan-centric repair can restore progress, where it fails, and which failure modes are most amenable to natural-language guidance, plan editing, multimodal annotation, or system-driven replanning.
    
\end{itemize}


%% file: 2_related_work.tex
\section{RELATED WORK}

\noindent
\textbf{Script-Based and Rule-Based GUI Automation}.
Traditional GUI automation systems, including commercial Robotic Process Automation (RPA) platforms, Selenium-style scripting frameworks, and record-and-replay tools, rely on low-level structural identifiers or fixed interaction traces to specify actions \cite{dwarakanath2018machines, nguyen2025gui, wewerka2020robotic, song2025can}. Although effective in stable and well-structured environments, these approaches are brittle: interface updates, layout shifts, and rendering changes can easily break scripts and require manual repair \cite{brisset2022erratum, yu2024practical, jain2024smartflow, yeh2009sikuli, coppola2019fragility}. As a result, they offer limited support for adaptation or recovery once execution deviates. 
Our work instead focuses on vision-based automation that operates over rendered interfaces while exposing task structure for user inspection and repair.

\vspace{0.5em}
\noindent
\textbf{Autonomous GUI Agents and Vision--Language Models}.
Recent advances in large language models (LLMs) and vision--language models (VLMs) have enabled GUI agents that operate directly over screenshots and natural language instructions, reframing automation as a perception--reasoning--action problem \cite{tang2025survey, qin2025ui, zhang2025btl}. Early systems demonstrated language-driven planning for web and mobile interfaces \cite{wen2024autodroid, song2024visiontasker}, while later work introduced stronger visual grounding and multimodal action generation that maps rendered pixels to executable actions \cite{shaw2023pixels, gou2024navigating, he2024webvoyager}. More recent foundation-level agents \cite{zhou2023webarena, koh2024visualwebarena, qin2025ui, wu2024atlas} and Vision-Language-Action architectures \cite{lin2025showui, yang2025ultracua, bhathal2025websight, huang2025spiritsight} have further expanded the scope of GUI automation and improved long-horizon execution across desktop and mobile environments. 
Many of these systems follow a ReAct-style paradigm \cite{yao2022react}, and recent work has begun to explore more explicit planning mechanisms for long-horizon GUI tasks \cite{ma2025agent+, erdogan2025plan}.

However, these systems remain largely autonomy-first. Planning and replanning are typically maintained as internal model state or exposed only as transient reasoning traces, giving users limited visibility into how actions are selected, how failures are interpreted, or how execution adapts over time \cite{hu2024dawn, xie2024osworld}. When execution drifts, recovery is usually handled by the agent alone, while user input remains confined to high-level natural language instructions rather than localized correction or collaborative revision of task structure. 
In contrast, our work externalizes evolving plans as persistent, editable, and inspectable artifacts, giving users direct leverage over how execution is reviewed, revised, and repaired over time.

\vspace{0.5em}
\noindent
\textbf{Interfaces for Human--Agent GUI Collaboration}.
Alongside increasingly capable GUI agents, recent work has begun to explore mixed-initiative and human-in-the-loop automation~\cite{amershi2019guidelines, vats2024survey, zou2025llm}. 
Early systems such as SUGILITE~\cite{li2017sugilite} showed that users can repair automation through multimodal programming-by-demonstration, and more recent work suggests that users often prefer ``do-it-with-me'' interaction over fully autonomous execution~\cite{khurana2025me}. 
Systems such as Cocoa~\cite{feng2024cocoa}, CowPilot~\cite{huq2025cowpilot}, and Magentic-UI~\cite{mozannar2025magentic} further support shared control through procedural co-authoring, redirected navigation, or editable task plans. 
Related frameworks also model users as active participants in agent workflows, emphasizing interaction-driven browsing and human-in-the-loop coordination~\cite{hua2025interactive, yun2025interaction, tang2026human}. 
These systems reintroduce user oversight into agent execution, but many still expose the agent's reasoning, recovery logic, or task structure only partially, making it harder for users to inspect, revise, and negotiate how execution evolves over time~\cite{chakraborti2019plan, rabanser2026towards}. Recent work such as DoubleAgents~\cite{long2025doubleagents} argues that plans should be treated as negotiable artifacts rather than fixed outputs. Rather than treating plan externalization as an end in itself, Plover uses the plan as the mechanism for repairing execution as it unfolds. Prior systems have shown that users can co-author, inspect, or edit agent plans, but Plover focuses on the moment when a GUI agent has already begun acting and its behavior starts to diverge from user intent. In this setting, the correction must specify what should change while preserving what has already been completed. Plover addresses this by making repairs operate on the remaining plan, grounded in the current screen state and recorded as part of the evolving workflow. This allows persistent plans to connect execution, correction, replanning, and provenance during GUI-agent repair, rather than serving only as previews, explanations, or editable task descriptions.


A related challenge is how users communicate situated corrections during GUI tasks. Because GUI interaction is inherently spatial, natural-language-only feedback is often insufficient in visually dense interfaces. Systems such as Morae~\cite{peng2025morae} and NaviPlus~\cite{cheng2025navi} address this challenge by surfacing ambiguity and asking clarifying questions during task execution. Other work explores deictic reference~\cite{maquil2023establishing, tsai2026uncertain} and multimodal grounding, including pointing, highlighting, freehand marks, and visual notes, as mechanisms for expressing intent or steering models~\cite{yen2025code, wu2024visual, wen2025exploring, shtedritski2023does, cai2024vip, chen2025interchat}. However, in most of these systems, such signals primarily serve as auxiliary inputs for disambiguation rather than as direct mechanisms for modifying automation. 
In contrast, our work treats visual annotations as actionable repair primitives over an explicit plan, enabling users not only to clarify intent but also to inspect, revise, and redirect execution as tasks drift. We also use annotation to resolve ambiguity, but the role of annotation in Plover is broader than disambiguating a target. Here, an annotation becomes an input to plan repair: it constrains which pending step or step sequence should be revised, anchors the revision to the current screen state, and produces an explicit proposal that the user can inspect before execution resumes. Thus, visual marks are not only grounding cues for action selection; they are repair primitives over a persistent task representation.

%% file: 3_sys_overview.tex
\section{PLOVER DESIGN}

\subsection{Problem Framing and Formative Study}

Autonomy-first GUI agents can often execute short tasks from natural language alone, but they remain difficult to supervise when execution unfolds over many steps in dynamic interfaces. In high-stakes settings, users must do more than specify a goal: they need to inspect evolving plans, detect breakdowns, and intervene without discarding prior progress.

Consider a compliance analyst using a legacy reporting portal to transfer values from a spreadsheet and PDF report into a multi-page form. She begins with a conventional autonomy-first GUI agent, which moves directly from instruction to action without exposing a stable plan she can inspect or revise (L1). Midway through execution, an unexpected modal dialog appears; the agent adapts, but does not reveal what changed or why (L2). Later, it enters a value into the wrong numeric field, yet the user has no precise, visually grounded way to redirect behavior at the point of failure (L3). When she attempts to correct the task through another prompt, the system broadly regenerates the workflow rather than modifying only the problematic portion (L4). Even when the agent reports completion, it does not clearly surface uncertainty, ambiguity, or intermediate failure (L5). This scenario is representative of workflows where users cannot fully ignore the automation outcome: the task spans multiple UI states, mistakes may silently propagate, and redoing the entire workflow is costly. In such cases, the relevant interaction problem is not whether the user should supervise every low-level action, but how the system can make the few necessary moments of supervision precise, localized, and recoverable. The scenario further illustrates five recurring limitations of autonomy-first GUI agents: \inlineLbox{L1} \textbf{Opaque Planning}, \inlineLbox{L2} \textbf{Silent Adaptation}, \inlineLbox{L3} \textbf{Weak Grounding for Correction}, \inlineLbox{L4} \textbf{All-or-Nothing Correction}, and \inlineLbox{L5} \textbf{Weak Verification}.

These limitations suggest that the challenge is not simply improving autonomous execution, but designing an interaction model that keeps task structure visible, supports targeted intervention, and preserves continuity across repair. To better understand these breakdowns, we conducted a formative study with six participants from our university community who had prior experience with agentic systems or GUI automation tools. Participants (P1--P6) completed a representative GUI task using an early prototype of \syss (Appendix Fig.~\ref{fig:old_interface}) while thinking aloud. Sessions lasted 30--60 minutes and covered prompt authoring, plan inspection and editing, execution monitoring, and multimodal correction of a deliberately introduced interface error. We collected screen recordings, verbal feedback, and post-task discussion. The study was deemed exempt by our University Institutional Review Board (IRB), and all participants provided informed consent.

Across participants, we observed four recurring challenges that motivated the design of \syss. \textbf{Planning Challenges:} users struggled to form accurate mental models of proposed plans and plan revisions before execution. \textbf{Execution and Inspection Challenges:} most participants struggled to judge whether execution was making meaningful progress, especially when repeated actions produced little visible change. \textbf{Intervention Challenges:} users wanted correction mechanisms that were precise, localized, and visually grounded; notably, all participants found screenshot annotation more precise than language alone for spatial correction, although several remained uncertain about how such inputs would affect the underlying plan. \textbf{System-level Challenges:} users expected the system to detect obvious non-progress and assist with recovery instead of placing the full burden of diagnosis on them. 

\subsection{Design Goals}
Our formative findings suggest that the main challenge in collaborative GUI automation is not only executing tasks correctly, but also helping users stay aligned with evolving execution. Across planning, execution, and intervention, participants repeatedly struggled to understand what the agent was about to do, determine whether it was making progress, and apply corrections without losing prior work. We synthesize these findings into three design questions: \textbf{Q1)} \textit{How can systems maintain alignment between evolving user intent and execution across multi-step workflows?} \textbf{Q2)} \textit{How can interfaces support precise, visually grounded correction when execution deviates? }\textbf{Q3)} \textit{How can replanning be surfaced as a transparent and inspectable process rather than a hidden internal update?} We then translate these questions into the following design goals:

\begin{enumerate}[label=\inlineGbox{\textbf{DG\arabic*}},leftmargin=*, itemsep=1pt, topsep=1pt]
    \item \textbf{Interpretable Plans.} Users should be able to understand, inspect, and revise plans before execution, with clear differentiation between plan versions and structure (Q1).
    \item \textbf{Execution Awareness.} The system should clearly communicate progress, failure, and state transitions during execution so that users can monitor behavior and recognize breakdowns (Q1, Q3).
    \item \textbf{Grounded Local Repair.} Users should be able to intervene through both high-level language and precise spatial grounding, while constraining updates to the relevant portion of the plan (Q2).
    \item \textbf{Failure-aware Recovery.} The system should identify non-progress and support recovery through explicit replanning rather than silent adaptation (Q3).
    \item \textbf{Continuity across Repair.} Intervention and replanning should preserve executed history and maintain a consistent sense of task progression (Q1, Q3).
\end{enumerate}

%% file: 4_plover.tex
\section{PLOVER SYSTEM}

\input{f1_system_framework}
\subsection{Overview}

\syss is a mixed-initiative system for plan-centric GUI automation. Its central design choice is to treat task plans as a persistent shared state representation \cite{horvitz1999principles} that coordinates planning, execution, and repair, rather than as transient internal reasoning. This choice addresses a key limitation of autonomy-first GUI agents: when plans evolve silently during execution, users have little visibility into what changed, why it changed, or how to intervene without restarting. By externalizing plans as persistent, editable artifacts \cite{feng2024cocoa, son2026hand}, \syss is designed to allow users to \textbf{Review} plan updates for alignment, \textbf{Revise} workflows through targeted intervention, and \textbf{Repair} failures without restarting execution.

As shown in Fig.~\ref{fig:system}, \syss separates planning from execution across three coordinated components: the \feature{Agentic Interface}, the \feature{Planner Service}, and the \feature{Executor Service}. A task begins when the user issues a prompt through the \feature{Agentic Interface} \circlep{1}, which forwards a \emph{Plan Request} to the \feature{Planner Service} \circlep{2}. The planner performs \mech{Plan Generation} to produce a versioned \emph{Structured Plan} that makes the intended task structure explicit. The plan is then passed to the \feature{Executor Service} \circlep{3}, which grounds each step into concrete keyboard, mouse, and observation actions within a \emph{Remote GUI Environment}. As execution proceeds, the executor reports \emph{Execution State} back to the interface \circlep{4}, which surfaces progress to the user through \emph{Execution Feedback} \circlep{5}.


This shared plan representation enables a coordination mechanism we call \maincontrib{Intelligent Replanning (IR)}: a visible plan-adaptation process that revises pending steps while preserving executed history. IR has two modes: \maincontrib{User-Driven IR}, triggered by user corrections, and \maincontrib{System-Driven IR}, triggered by non-progress detection. In \maincontrib{User-Driven IR}, users intervene upon observing an error through \mech{Plan Edits}, \intervention{Natural Language Guidance}, or \intervention{Multimodal Annotation}, \circlep{1} triggering a \emph{Replan Request} \circlep{2} that revises only pending steps while preserving executed history. Conversely, in \maincontrib{System-Driven IR}, the executor autonomously detects non-progress via \mech{Non-progress Detection} \circleg{1}, prompting the planner to propose a localized recovery step surfaced explicitly on the interface \circleg{2}. In both modes, revisions update the shared plan, allowing users to inspect the rationale and scope of changes. This coordination loop transforms GUI automation from a brittle one-shot interaction into a collaborative, repairable process.


\subsection{Agentic Interface}
\input{f2_interface}

The \feature{Agentic Interface} is the primary interaction surface through which users inspect plans, monitor execution, and steer adaptation. Fig.~\ref{fig:interface} shows a typical interaction: users author a task (b), review and optionally revise the generated plan (c), monitor execution through screen state and step-level feedback (d, e), and intervene through \intervention{Natural Language Guidance} (b) or \intervention{Multimodal Annotation} (e1) when execution drifts. Resulting IR events are versioned (f) so that users can inspect how the workflow changes over time while preserving completed progress. Rather than treating planning, execution, and replanning as separate stages, the interface presents them as a continuous workflow over a shared plan representation. 
It therefore emphasizes three primary functions: to \textbf{Review} plan alignment, \textbf{Revise} steps via intervention, and \textbf{Repair} failures.

\paragraph{Shared Plan Workspace}
To address opaque planning and silent adaptation (\inlineLbox{L1}, \inlineLbox{L2}), \syss externalizes task structure in a shared workspace centered on the \feature{Planner Chat} (Fig.~\ref{fig:interface}b) and \feature{Steps Panel} (Fig.~\ref{fig:interface}c). Plans are represented as editable \mech{Plan Artifacts} that users can inspect, revise, and approve before and during execution in the Regenerated Plan Panel (Fig. \ref{fig:interface}c2) (\inlineGbox{DG1}). This workspace is not merely a display of model output; it serves as the primary coordination object between the user and the agent. After the user specifies a task in the chat box, the planner generates a structured step sequence in the Steps Panel. Before execution, users may regenerate alternatives and approve a preferred version to enter a locked execution state. During execution, revisions introduced through \intervention{Natural Language Guidance} or \intervention{Multimodal Annotation} are surfaced as explicit proposals with localized changes (\inlineLbox{L3}) rather than silently overwriting the active plan. By separating the current plan from proposed revisions, \syss provides a mechanism for users to evaluate repair before committing to it, reducing the all-or-nothing correction behavior (\inlineLbox{L4}).
To further support interpretability, each replanning event updates \feature{Derived Constraints} (Fig.~\ref{fig:interface}c1), shown as a concise summary of the agent's inferred understanding of user intent. A dedicated proposal view supports side-by-side comparison between the current and proposed plans and includes a concise diff summary of updated steps (Fig.~\ref{fig:annotation_flow}d). Together, these mechanisms are designed to make revision reviewable and support continuity across repair (\inlineGbox{DG5}).

\paragraph{Execution Legibility}
To support execution awareness without overwhelming users (\inlineGbox{DG2}), the \feature{Executor Panel} (Fig.~\ref{fig:interface}e) presents agent behavior through a step-centered execution view. It combines a live screenshot panel (e1), a concise semantic activity feed (e2), and a lightweight step timeline (e3).
The screenshot panel shows the current interface state and supports direct \intervention{Multimodal Annotation} when correction is needed (Fig.~\ref{fig:annotation_flow}b). The activity feed summarizes low-level operations such as clicking, typing, scrolling, and waiting into short human-readable descriptions that foreground agent intent rather than implementation detail. The step timeline shows which steps are completed, active, or failed. This design reflects a deliberate tradeoff: users need enough visibility to judge whether execution is making progress, but raw tool traces alone would impose unnecessary cognitive load. \syss therefore uses progressive disclosure (\inlineLbox{L2}, \inlineLbox{L5}): high-level execution summaries are visible by default, while a secondary drawer (Fig.~\ref{fig:interface}e4) supports deeper inspection through detailed logs and reasoning traces.

\paragraph{System Status Bar}
To reduce ambiguity about current system behavior (\inlineLbox{L2}), \syss provides a persistent \feature{System Status Bar} (Fig.~\ref{fig:interface}a) that communicates execution state and plan state at a glance. It indicates whether the system is planning, replanning, executing, paused, failed, or waiting for user input, while separately representing the plan as \emph{Draft}, \emph{Proposal}, or \emph{Locked}. The bar also surfaces step progress, current grounding context, connectivity to the execution environment, elapsed time, and execution mode. This persistent overview gives users a stable frame of reference during mixed-initiative interaction and supports both execution awareness and failure recovery (\inlineGbox{DG2}, \inlineGbox{DG4}).

\paragraph{Run Timeline and Provenance}
\input{f3_provenance}

To support mental model continuity and transparent plan evolution (\inlineGbox{DG2}, \inlineGbox{DG5}), \syss visualizes execution and revision history through a two-level iconographic \feature{Provenance Component} (Fig.~\ref{fig:interface}f). This design addresses formative study findings where participants struggled to track plan changes and understand how interventions affected ongoing execution.
A persistent Run Timeline (Fig. \ref{fig:interface}f) represents the active plan as a sequence of status-based icon nodes corresponding to pending, executing, completed, failed, and IR recovery states. Small adjacent histograms \includegraphics[height=1em]{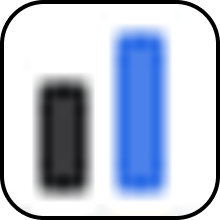} summarize execution metrics (e.g., latency and screenshot counts), while between-node intervention markers denote user interventions through \includegraphics[height=1.1em]{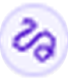} \intervention{Multimodal Annotation} or \includegraphics[height=1.1em]{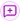} \intervention{Natural Language Guidance}. Expanding the timeline reveals a branching version graph of approved plan revisions \cite{chakraborti2017visualizations, narechania2025utilizing} (Fig. \ref{fig:interface}f1; Fig. \ref{fig:provenanance_bar}). Each row corresponds to a version labeled by revision cause (e.g., initial plan, manual regeneration, instruction update, annotation update, or \maincontrib{Intelligent Replanning}), with edges connecting revisions to their origin points (similar to a Git-style branching history). 
This provenance view serves a broader purpose than history logging. Since mixed-initiative automation unfolds through successive revisions, users need to understand not only the current plan, but also how the workflow arrived there. By making plan evolution explicit, \syss turns replanning from a hidden recovery mechanism into an auditable interaction process (\inlineLbox{L2}, \inlineLbox{L5}).

\subsection{Intelligent Replanning} A central mechanism exposed through the interface is \maincontrib{Intelligent Replanning (IR)} (Fig. \ref{fig:system}), which treats plan adaptation as a visible and revisable process (\inlineLbox{L1}). \maincontrib{IR} allows users to \textbf{Review} plan updates and \textbf{Revise} workflows through \maincontrib{User-Driven IR}, while \textbf{Repairing} failures through \maincontrib{System-Driven IR}, in which the system detects non-progress and proposes localized recovery steps (\inlineGbox{DG4}). Both operate over the same shared plan representation, ensuring that updates remain inspectable, versioned, and minimally disruptive.

\paragraph{User-Driven IR}

\maincontrib{User-Driven IR} allows users to intervene during execution by providing corrective signals grounded in either language or visual context. These interventions generate localized updates that \textbf{modify only pending steps} while preserving executed history (\inlineGbox{DG3}, \inlineGbox{DG5}). The key design principle is that user feedback should not trigger monolithic regeneration unless the task itself has fundamentally changed.

\textbf{(1) \intervention{Natural Language Guidance}.}
Users may provide free-form instructions to revise execution, such as ``select the second option instead'' or ``skip this step and go to export.'' These inputs are interpreted as high-level intent corrections and translated into localized edits of the relevant plan segment. This provides flexibility and low interaction overhead, but it can remain ambiguous in visually dense interfaces where multiple candidate targets are plausible.

\textbf{(2) \intervention{Multimodal Annotation}.} 
To resolve ambiguity, \syss supports screenshot-based annotation (Fig.~\ref{fig:annotation_flow}b).
When the execution feed reveals execution drift  ({Fig.~\ref{fig:annotation_flow}a}), users can mark the screenshot using strokes, shapes, or text overlays, producing a multimodal artifact anchored in pixel space.
The UI maintains annotation primitives $\mathcal{A}=\{a_k\}$ and computes a bounding box $b(\mathcal{A})=(x_{\min}, y_{\min}, w, h)$ over all primitives. On save, the artifact consisting of the annotated screenshot $I_t$ and geometry $b(\mathcal{A})$, is submitted to the \feature{Planner Service}, with confirmation via status banners in the Planner Chat ({Fig.~\ref{fig:annotation_flow}c}).
In \syss, this annotation constrains repair \emph{scope} by prioritizing spatial signals over language. The resulting schema-constrained request triggers localized modifications that preserve executed history, surfaced as a revised plan proposal grounded in the annotated region ({Fig.~\ref{fig:annotation_flow}d}). Once the user approves the proposal, the \feature{Executor} acts on the specified target ({Fig.~\ref{fig:annotation_flow}e}). By favoring the smallest consistent update, \maincontrib{User-Driven IR} ensures precise corrections and execution continuity, with the entire intervention recorded as a persistent marker in the \feature{Run Timeline} ({Fig.~\ref{fig:annotation_flow}f}).

\input{f4_annotation_flow}


\paragraph{System-Driven IR}

In addition to explicit user intervention, \syss triggers \maincontrib{System-Driven IR} when execution appears to stall. Autonomous GUI agents often fail through repeated actions that produce little meaningful interface change. Detecting such failures reliably is non-trivial: repeated actions alone may reflect benign retries, while a visually static interface may simply indicate legitimate waiting. Inspired by works like \cite{hao2025uncertainty, wu2025gui}, \syss therefore combines behavioral and visual signals to detect non-progress conservatively.

\textbf{(1) Behavioral Loop Detection.}
The executor maintains a canonicalized sequence of recent actions $\mathcal{H}_t = (a_1, a_2, \dots, a_t)$, where actions are abstracted to coarse semantic types (e.g., click, scroll, wait). Repeated subsequences beyond a threshold indicate potential behavioral thrashing.

\textbf{(2) Visual Non-Progress Verification.}
To determine whether these repeated actions changed the environment, the system evaluates perceptual changes in the interface.
For each screenshot $s_t$, a perceptual hash $h_t = \mathrm{dHash}(s_t)$ is computed, and similarity between consecutive states is measured using Hamming distance $d_t = \mathrm{Hamming}(h_t, h_{t-1})$. If $d_t < \tau$ over multiple steps, the interface is considered visually stable, indicating a likely lack of progress.

\textbf{(3) Recovery Policy.}
Execution is classified as \emph{stuck} only when both behavioral repetition and visual stability are observed, reducing false positives from ordinary retries or transient waits. When triggered, the system halts the current step, marks it as \emph{Failed}, and inserts a corrective step as an IR recovery step. The proposed recovery is surfaced with a short rationale so users can understand why replanning occurred and what change is proposed.

\subsection{Planner}

The \feature{Planner Service} synthesizes and revises structured plans that mediate between user intent and GUI execution. Rather than regenerating behavior monolithically, \syss treats plans as persistent, reviewable artifacts whose executed history is preserved while remaining steps can be selectively revised, enabling interpretable planning and localized repair (\inlineGbox{DG1}, \inlineGbox{DG3}, \inlineGbox{DG5}).

\paragraph{Architecture.}
The planner mediates communication between the interface, the language model, and the executor. On each invocation, the interface submits a structured interaction context and receives a schema-constrained plan update. The planner does not directly control the execution environment. Instead, the environment state is supplied through a serialized interaction context, including user instructions, plan history, and optional screenshot metadata. Initial planning operates over task instructions and conversation history, while replanning with \intervention{Multimodal Annotation} additionally incorporates the latest screenshot and annotation geometry. 
This separation is deliberate. Decoupling planning from direct actuation keeps plan updates reviewable at the representation level rather than allowing hidden policy shifts inside the executor.
To manage token constraints, the service retains full plan history while limiting visual context to the three most recent screenshots. This structured update mechanism enables visible adaptation rather than silent internal changes (\inlineLbox{L1}, \inlineLbox{L2}).

\paragraph{Plan Representation and State Invariants.}


The planner outputs responses using a two-block schema. The \texttt{<analysis>} block summarizes task context, while the \texttt{<steps>} block contains ordered \texttt{<completed>} and \texttt{<pending>} sequences. 
Completed steps represent immutable execution history, while pending steps define the editable remainder. Each step is expressed as a deterministic UI instruction to support direct grounding by the executor. We formalize the planner state at time $t$ as $P_t=(C_t, U_t)$, where $C_t$ denotes executed steps and $U_t$ the editable suffix. Given input $X_t$, the planner computes $P_{t+1}=\mathrm{Update}(P_t,X_t)$ subject to the invariant $C_{t+1}=C_t$. 
This invariant captures a core design principle in \syss: once a step has been executed and committed to shared history, repair should operate over the remaining suffix rather than silently rewriting the entire workflow. In practice, this constraint makes revisions localized, preserves provenance, and prevents the broad regeneration behavior (\inlineGbox{DG3}, \inlineLbox{L4}) that users found difficult to interpret in autonomy-first systems.
When replanning is triggered either through user intervention or system detection, the planner applies schema-constrained updates conditioned on the current plan $P$ and available context, such as annotation artifacts $(I_t, b(\mathcal{A}))$. These updates occur at three granularities: a \mech{local edit} (single-step modification), a \mech{regional patch} (short sequence update), or a \mech{full replan} (workflow restructuring). This hierarchy enables the planner to adapt flexibly while preserving plan continuity and interpretability.

\subsection{Executor}
The \feature{Executor Service} translates structured plan steps into concrete GUI interactions while maintaining a strict separation between reasoning and actuation (\inlineGbox{DG2}, \inlineGbox{DG4}).
Each executor implements a shared RPC contract exposing deterministic primitives---pointer actions, keyboard input, scrolling, waits, and observation endpoints such as screenshots. This abstraction serves two purposes. First, it enforces a clear boundary between high-level planning decisions and low-level execution, ensuring that task logic remains visible in the plan representation rather than being hidden inside execution heuristics \cite{zhang2026showui, aghzal2026llm}. Second, it supports portability across heterogeneous GUI environments by keeping the planner interface stable even when the underlying actuation layer changes (see Appendix \ref{sec:appendix_implementation}).
To support mixed-initiative control, \syss also provides a parallel human-access channel to the execution environment. However, the system prioritizes lightweight, plan-centric corrections, such as \intervention{Natural Language Guidance} and \intervention{Multimodal Annotation}, over full manual override, because these interventions preserve explicit task structure and provenance. Currently, manual interventions can alter execution behavior but are not yet incorporated back into the plan representation; integrating explicit plan updates for human overrides remains future work.


%% file: f1_system_framework.tex
\begin{figure}[t]
  \centering
  \includegraphics[width=0.48\textwidth]{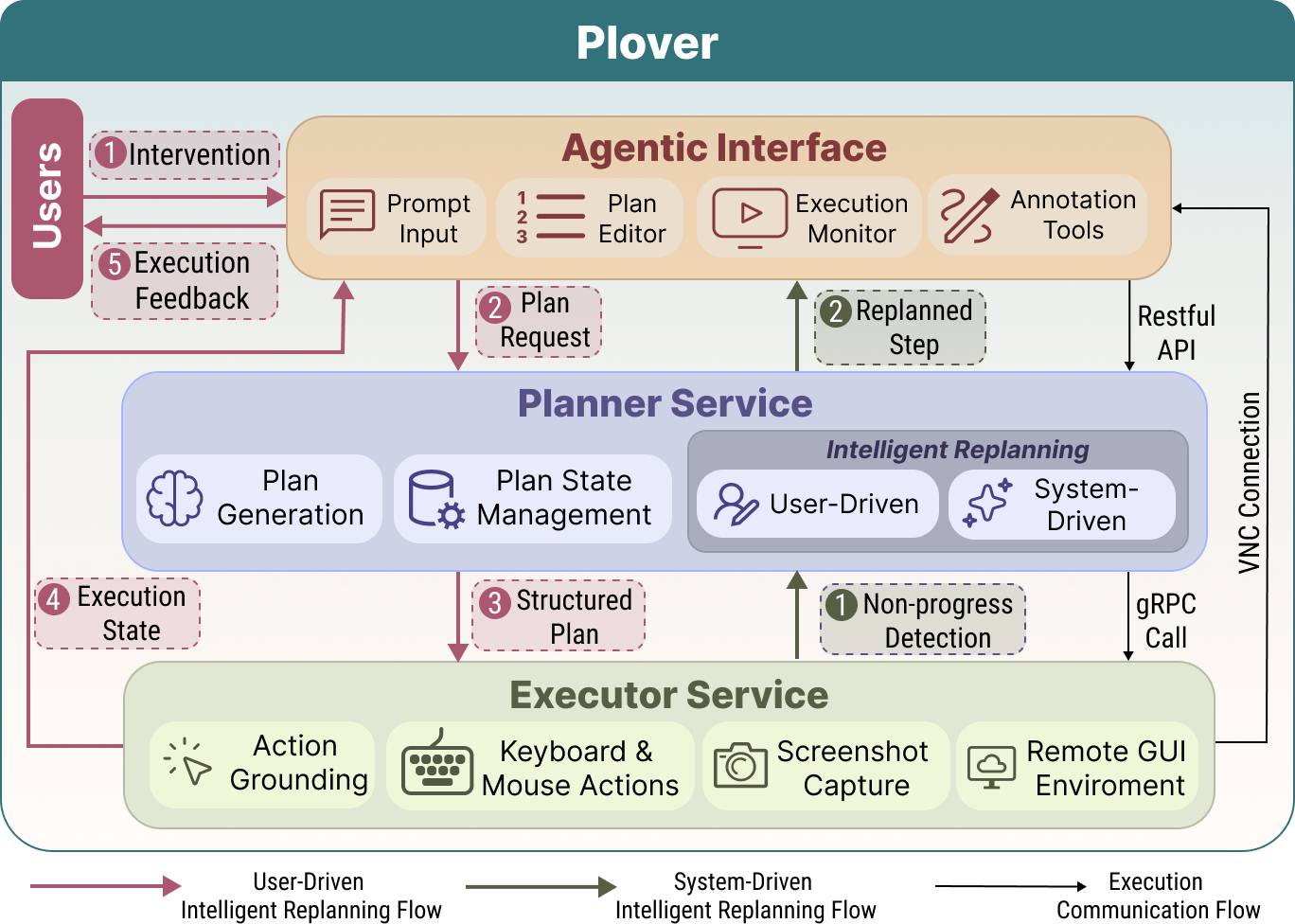}
   \caption{The \textsc{Plover} system architecture. }\Description{Diagram of the Plover system architecture showing its main components and flows.}
  \label{fig:system}
  \vspace{-2.5em}
\end{figure}

%% file: f2_interface.tex
\begin{figure*}[t]
  \centering
  \includegraphics[width=\textwidth]{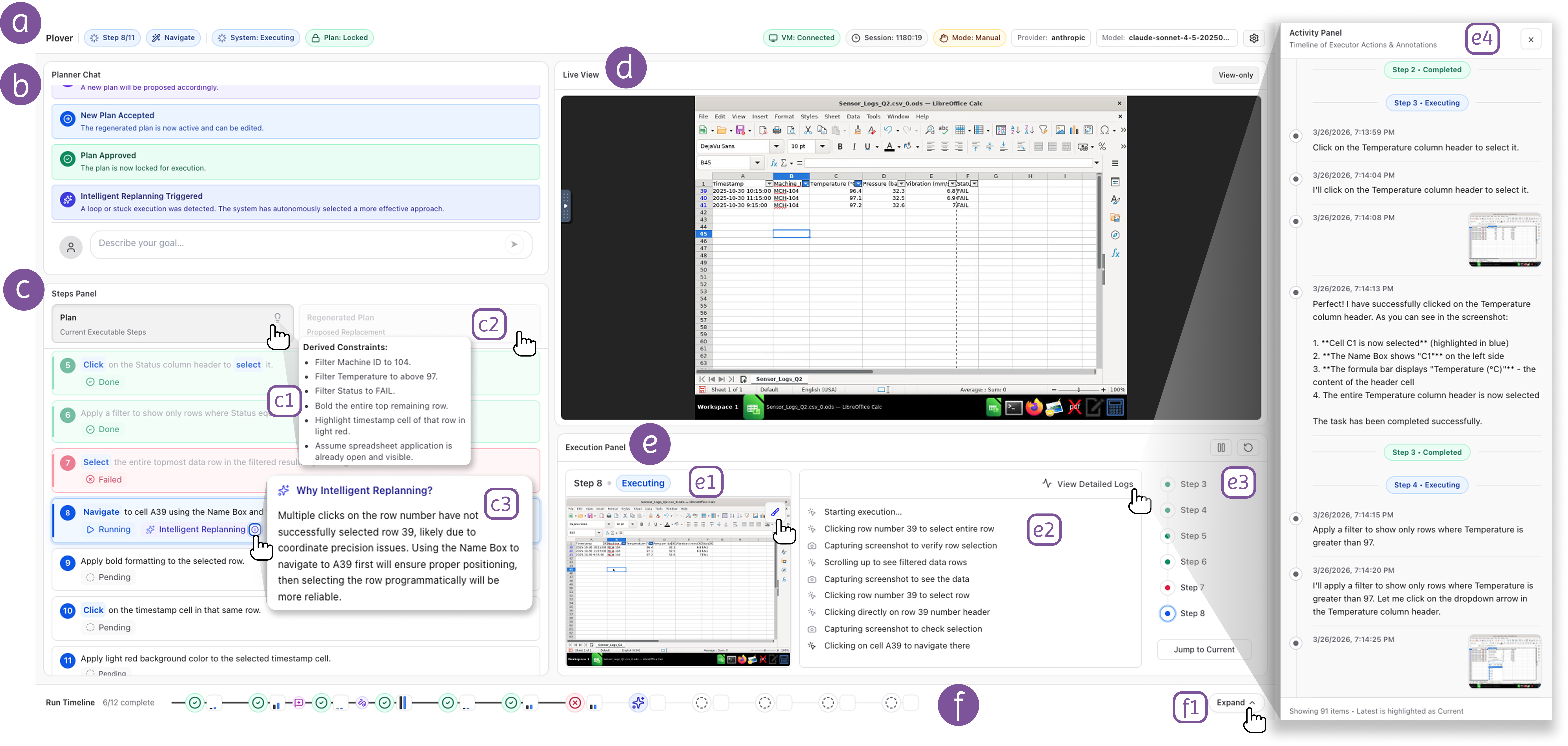}
  \Description{Screenshot of the Plover Agentic Interface.}
  \caption{Plover Agentic Interface. (a) System Status Bar shows execution phase and plan state. (b) Planner Chat supports natural language interaction and surfaces intervention updates as banners. (c) Steps Panel displays the interactive plan, Derived Constraints (c1), Regenerated Plan Proposals (c2), and \maincontrib{System-Driven IR} proposals (c3). (d) Live View provides visual grounding of the target environment. (e) Execution Panel includes annotation tools (e1), execution feed (e2), step timeline (e3), and expanded activity logs (e4). (f) Run Timeline provides an iconographic view of execution progress and access to expanded branching plan visualization (f1; Fig. 3).}
  \label{fig:interface}
  \vspace{-1em}
\end{figure*}

%% file: f3_provenance.tex
\begin{figure}[t]
  \centering
  \includegraphics[width=0.48\textwidth, trim=1 1 1 1, clip]{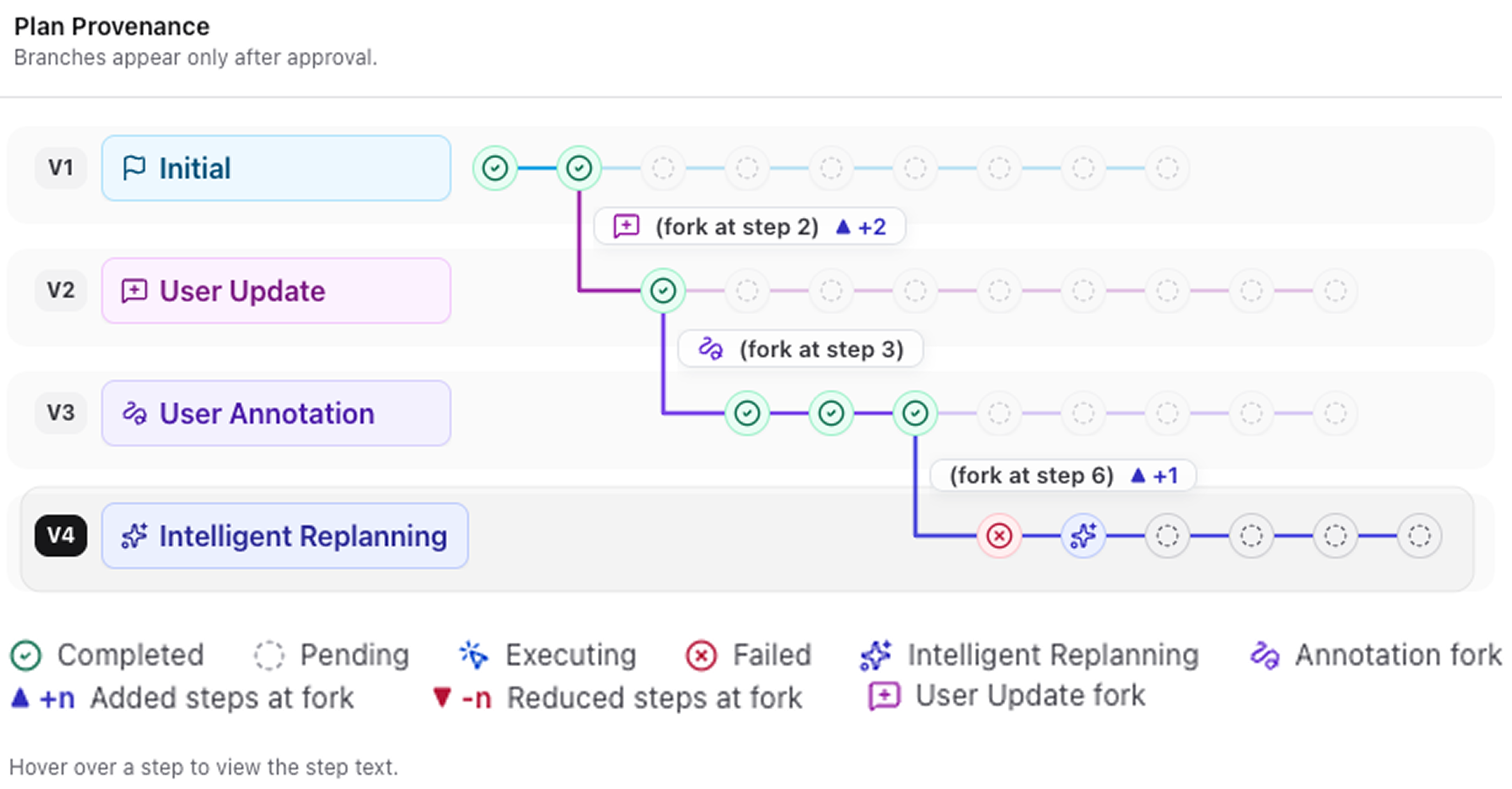}
  \caption{Provenance Bar visualizing branching plan revisions and evolution.}
  \label{fig:provenanance_bar}
  \vspace{-2em}
\end{figure}

%% file: f4_annotation_flow.tex
\begin{figure}[t]
  \centering
  \includegraphics[width=0.48\textwidth]{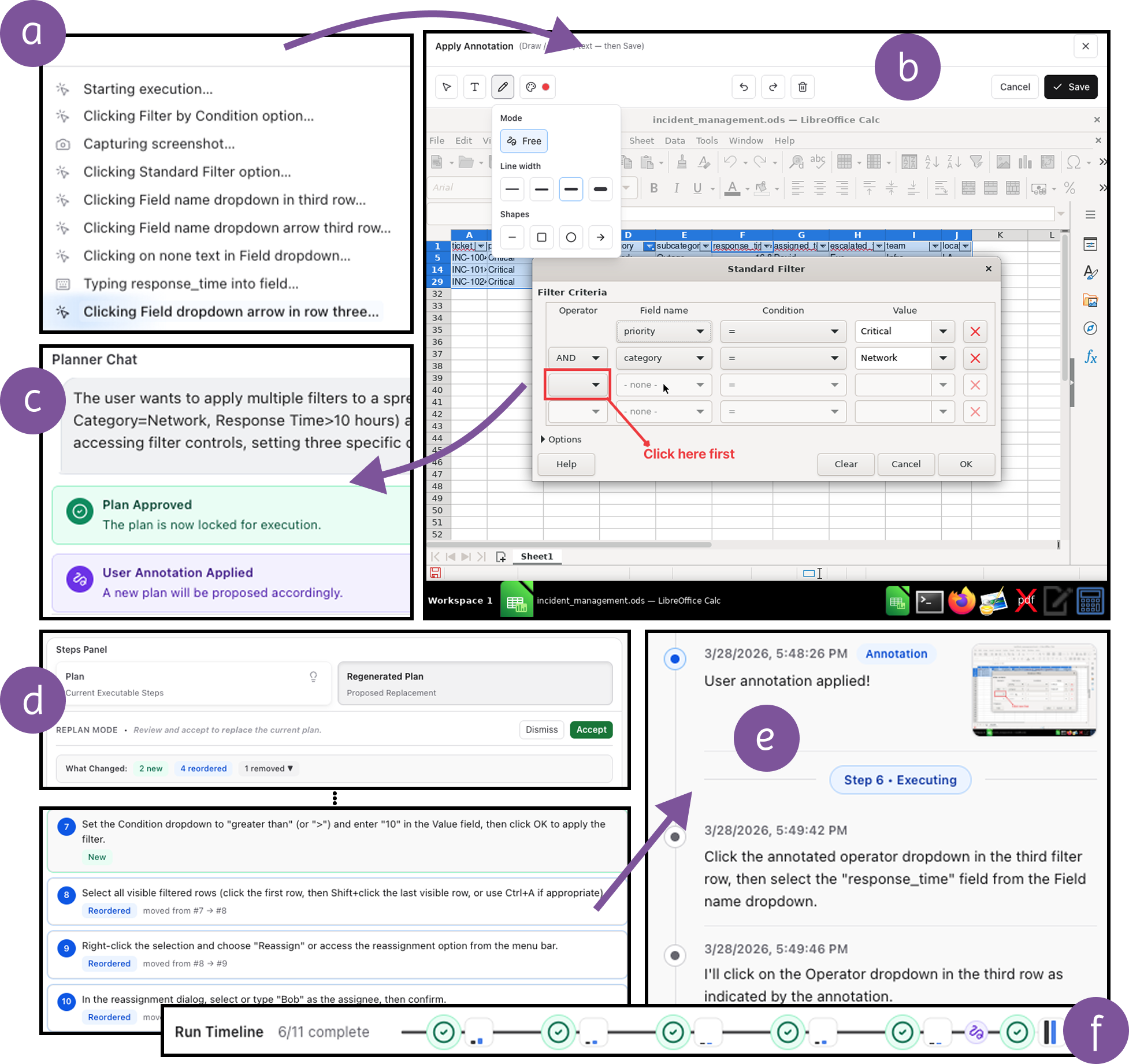}
  \caption{Multimodal Annotation workflow for resolving spatial ambiguity through localized, plan-centric repair.}
  \label{fig:annotation_flow}
  \vspace{-2em}
\end{figure}

%% file: 5_evaluation.tex
\section{EVALUATION}
\input{t1_failure_cases_summary}

We evaluate \syss as a system for recoverable automation through three complementary analyses of failure and repair: (1) whether autonomous agent failures are structurally recoverable under plan-centric mixed-initiative interaction, (2) how stable autonomous execution and plan reconstruction remain in realistic multi-step workflows, and (3) which classes of failures are most amenable to repair. Rather than focusing only on end-to-end success rates, our evaluation tests \sys’s core claims: that \textbf{many GUI-agent failures are structurally recoverable, exposing plans makes deviations easier to diagnose, and lightweight interventions can restore progress}.
The benchmark repair study (Section~5.1) measures recoverability under mixed-initiative interaction. The scenario-based stability study (Section~5.2) examines autonomous execution and plan alignment in realistic workflows. We then synthesize both through a failure taxonomy identifying when different repair mechanisms are most effective (Section~5.3). Together, these analyses show whether \syss improves outcomes, and how plan-centric interaction reshapes GUI automation dynamics.


\subsection{Benchmark Failure-Case Repair Analysis}
\label{sec:benchmark_failure}

\input{t1_eval_summary_osworld}

We first evaluate the central claim of \syss: that many autonomous GUI-agent failures are recoverable when plans are externalized and repair is localized. To do so, we curated 38 tasks from the OSWorld-Verified benchmark that were previously reported as failures for Claude 4.5 Sonnet using native computer-use capabilities \cite{xie2024osworld}. OSWorld-Verified is a curated subset of OSWorld with verified task definitions, initial states, and evaluation procedures. These tasks, detailed in Appendix \ref{sec:appendix_evaluation}, span heterogeneous environments, including web browsers and desktop applications such as LibreOffice, and emphasize multi-step workflows known to challenge vision-based automation \cite{drouin2024workarena, boisvert2024workarena++, nayak2025ui}. We first ran all tasks in a fully autonomous setting to establish a baseline. Of the 38 tasks, 26 resulted in non-success outcomes (partial or failure); these cases formed the basis for our repair analysis.

We then re-ran the 26 autonomous non-success cases in a mixed-initiative setting, where an expert (first author) provided targeted interventions---richer than mere prompt re-issuance---using \intervention{Natural Language Guidance}, \intervention{Multimodal Annotation}, and \mech{Plan Edits}, while \syss triggered \maincontrib{System-Driven IR} when drift was detected. Interventions were lightweight and localized: natural-language guidance typically used 1--2 sentences clarifying spatial references or correcting element selection (e.g., ``\textit{use the supplier dropdown on the left, not the archived list}'')($<$15s), multimodal annotations marked specific screen regions and sometimes added a short text label ($<$30s), and plan edits involved editing, adding, removing, or reordering steps in the initial or pending plan($<$20s). No intervention required code, system internals, or domain knowledge beyond what was visible in the interface. \textbf{This setup establishes an upper bound on plan-centric recoverability rather than typical user performance}. That choice is intentional: our goal is to test whether failures are \textit{structurally recoverable} through plan-centric interaction, not how quickly naive users can identify and apply corrections. Establishing recoverability as a property of the failure class is a prerequisite for later user-facing studies \cite{potts2026invisible}. 

Mixed-initiative interaction improved 23 of 26 autonomous non-success cases, converting 17 to complete success and 6 to partial success (completing subtasks but not the goal), with only 3 remaining failures (Table~\ref{tab:osworld_summary}). Recovery required an average of 2.04 interventions per task. Notably, all 10 autonomous partial successes were converted to complete success, and no regressions were observed. Improvements were consistent across application types, with 100\% recovery in browser and writer tasks, and substantial gains in Calc (86\%) and multi-application workflows (80\%). Together, these findings suggest that many GUI-agent failures are not irrecoverable planning errors, but localized breakdowns in grounding, state interpretation, or execution continuity that become structurally recoverable when plans are externalized and repair is localized.


\subsection{Scenario-Based Stability Analysis}
\input{t2_eval_image_plan_combined}
We next examine the complementary setting in which no user intervenes. Whereas the repair study tests recoverability, this analysis probes how well \syss maintains execution stability and plan alignment in realistic multi-step workflows. To do so, we constructed five practical scenarios involving dense interaction patterns such as constrained typing, multi-step filtering, and application navigation. Rather than manually authoring prompts, we generated realistic task instructions from sampled interaction trajectories and replayed them through \syss without human intervention. This setup tests whether \syss can reconstruct plausible task structure and maintain execution alignment in autonomy-only workflows. Scenario details are provided in Appendix \ref{sec:appendix_evaluation}.

\paragraph{Trajectory Sampling and Prompt Reconstruction}
To generate diverse task instances, we built a constrained exploration pipeline sampling plausible interaction trajectories within each scenario, following established methods for automated GUI data collection \cite{guo2025auto}. For each environment, we specify initialization commands, curated clickable regions, restricted regions, and typing regions paired with text inputs. These constraints define valid interaction spaces over screenshots. During exploration, the system captures screenshots, queries a vision-based inference endpoint with Omniparser \cite{lu2024omniparserpurevisionbased} for candidate actions, and executes only actions allowed by the scenario specification. We ran 100 exploration trials per scenario with randomized trajectory lengths of 5--15 steps, then filtered out redundant traces and retained 5 diverse trajectories per scenario after manual verification, yielding 25 trajectories overall \cite{kang2026learning}.
We convert each retained trajectory into a natural-language task description using Claude Sonnet 4.5, employing a reverse task synthesis approach to infer user intent \cite{sun2025genesis}. 
We then replay each generated prompt through the full \syss pipeline without human intervention. This setup tests whether \syss can reconstruct a plausible plan and reach a comparable end state from prompts inferred from behavior rather than manually authored instructions.

\paragraph{Visual Fidelity}

Autonomous replay proved substantially more stable in browser workflows than in desktop workflows. To assess replay fidelity, we compare the final interface state produced during replay against the reference trajectory using structural similarity (SSIM) \cite{wang2004image}, mean squared error (MSE), and perceptual hashing (dHash) \cite{madden2024robustness}. As shown in Table~\ref{tab:combined_eval_summary}, Firefox-based scenarios achieve consistently high similarity (SSIM $>$ 0.98), whereas LibreOffice scenarios exhibit substantially lower similarity (SSIM 0.61--0.69) and higher variation (Appendix \ref{sec:appendix_evaluation}). We treat image similarity as a proxy for execution stability rather than task success. These results suggest that browser workflows are relatively stable and reproducible, while desktop workflows are more sensitive to intermediate state differences, layout changes, and modal interactions. \textbf{This gap highlights when autonomy breaks down and why plan-centric repair becomes necessary, particularly in desktop settings.}

\paragraph{Plan Alignment}
Autonomously generated plans were often executable but structurally imperfect. We compare generated plans against reference trajectories using phase coverage, order alignment, redundancy, and actionability, following the Goal-Plan-Action (GPA) alignment framework \cite{jia2025your}. To enable comparison, both plans and trajectories are mapped to a shared interface-agnostic phase taxonomy (e.g., navigation, filtering, selection), collapsing consecutive duplicates. Coverage measures recovered phases, order measures sequence agreement, redundancy captures repeated phases \cite{grigorev2025verifyllm}, and actionability measures executable steps.

As shown in Table~\ref{tab:combined_eval_summary}, actionability remains high (0.97), while coverage (0.62) and order alignment (0.41) are only moderate. This gap between step-level executability and workflow-level alignment \cite{balepur2025good} is exactly the space that \syss targets: \textbf{plans may remain locally plausible even when they are globally misaligned, creating an opportunity for users to inspect and correct them before deviations propagate}. 
Non-trivial redundancy (0.33) further suggests that predicted plans often reach subgoals through reordered or repetitive paths, which in autonomy-only settings can accumulate into drift or partial completion \cite{grigorev2025verifyllm, wei2025plangenllms}. We also observe that higher redundancy correlates with lower replay fidelity, suggesting that \textbf{repeated phases often reflect unsuccessful recovery attempts}. Trial-level results are reported in Appendix \ref{sec:appendix_evaluation}.

\subsection{Characterizing Repairable Failures}
To understand when mixed-initiative interaction is most effective, we analyzed the 26 autonomous non-success cases and examined how different interventions restored progress. Fig. \ref{fig:failure_recovery} summarizes representative failure archetypes and their corresponding repair pathways. A key observation emerges: \textbf{repairability is strongly tied to the locality of the failure}. Breakdowns such as execution drift, perception errors, planning errors, and state misinterpretation (Fig.\ref{fig:failure_recovery}a-d) remain recoverable through targeted, localized interventions. In contrast, compound failures (Fig.~\ref{fig:failure_recovery}e), where multiple errors propagate across dependent steps, proved significantly harder to repair. Here, failure was no longer localized to a single action or state, but distributed across the plan, execution history, and current interface context. Under these conditions, each repair intervention provided only partial corrective leverage, since none could fully reconstruct the lost structural alignment of the task. \textbf{This suggests that plan-centric interaction is most effective for localized breakdowns, but becomes fundamentally constrained when early errors cascade into globally misaligned workflows.}

\paragraph{Planning Errors and Execution Drift.} 

\textit{Execution Drift} emerged as the primary failure mode, appearing in 46\% ($n=12/26$) of cases. One-third of these instances ($n=4$) were \textit{Compound Failures}, where drift stemmed from underlying \textit{Planning Errors}. These compound cases were the most difficult, accounting for two of the only three unrecovered tasks ($n=3/26$). Despite this complexity, \syss effectively arrested divergent behavior in the remaining instances. \intervention{Natural Language Guidance} was the most frequent recovery channel (resolving 11 cases, 5 exclusively), while \maincontrib{System-Driven IR} and \intervention{Multimodal Annotation} each successfully resolved 9 cases (3 each exclusively). These findings suggest that while compound errors represent the current boundary of agent autonomy, \syss's key contribution is that \textbf{explicit, localized replanning can restore progress before drift becomes terminal.}


\paragraph{Perception and Grounding Breakdowns.} 
\textit{Perception Errors} ($n=7/26$) and \textit{Action Grounding Failures} ($n=5$) occurred when agents selected incorrect UI elements despite correct intent. These failures were particularly common in visually dense interfaces such as LibreOffice and multi-application workflows. Most were recoverable through \intervention{Natural Language Guidance} or \intervention{Multimodal Annotation}, which provided contextual clarification or spatial grounding. Replay analysis further showed lower visual fidelity in LibreOffice scenarios (SSIM as low as 0.61), reflecting increased visual ambiguity and sensitivity to intermediate state differences. 

\paragraph{State Misinterpretation and Missing Context.} 
\textit{State Misinterpretation} ($n=5/26$) and \textit{Missing Context} ($n=2$) occurred when agents misinterpreted modal dialogs, document states, or required domain-specific inputs. These failures were mostly resolved through brief \intervention{Natural Language Guidance}, suggesting that 
\textbf{many breakdowns stemmed from incomplete local context rather than fundamentally incorrect task decomposition.}

\paragraph{Scenario-Based Patterns}
Failure patterns also varied across domains. Browser tasks achieved full recovery across all non-success cases, while multi-application workflows were the most challenging, accounting for 2 of the 3 remaining failures and requiring the highest number of interventions. 
In these harder cases, the initial planning error produced a fundamentally misaligned decomposition that propagated through dependent steps. Even after user- and system-driven interventions, the accumulated divergence was too large to recover through localized edits alone, suggesting that some failure modes require earlier intervention during planning rather than mid-execution repair. 
 Replay metrics support the same pattern, with browser scenarios achieving higher visual similarity (SSIM $\approx$ 0.98) compared to LibreOffice scenarios (SSIM 0.61–0.69). Despite these differences, high actionability (0.97) suggests that plans were technically sound, though valid planning alone did not guarantee successful execution grounding when alignment was imperfect. 

Overall, failures were typically incremental and locally repairable. Mixed-initiative interaction improved 88\% of non-success cases with an average of 2.04 interventions per task, and most improvements occurred after lightweight, localized corrections. Ultimately, these results indicate that \textbf{GUI agent reliability depends less on perfect autonomy and more on providing the plan-centric visibility required to transform latent drift into a collaborative and repairable interaction.}

%% file: t1_failure_cases_summary.tex
\begin{figure*}[t]
\centering
\small
\bfseries
\setlength{\tabcolsep}{3pt} 
\renewcommand{\arraystretch}{1.15}
{
\arrayrulecolor{gridclr}
\rowcolors{2}{stripebg}{white}

\begin{tabular}{|c|
>{\raggedright\arraybackslash}p{3.8cm}|
>{\raggedright\arraybackslash}p{1.9cm}|
c|
>{\centering\arraybackslash}m{2.5cm}|   
>{\raggedright\arraybackslash}p{4cm}|
>{\raggedright\arraybackslash}p{1.9cm}|
c|}
\hline
\rowcolor{headerbg}
\textbf{} & \textbf{Auto State} & \textbf{What Went Wrong} & \textbf{Auto} & \textbf{Failure Type} & \textbf{Intervention} & \textbf{How It Was Repaired} & \textbf{MI} \\
\hline

(a) &
\raisebox{-0.9\height}{\includegraphics[width=3.8cm]{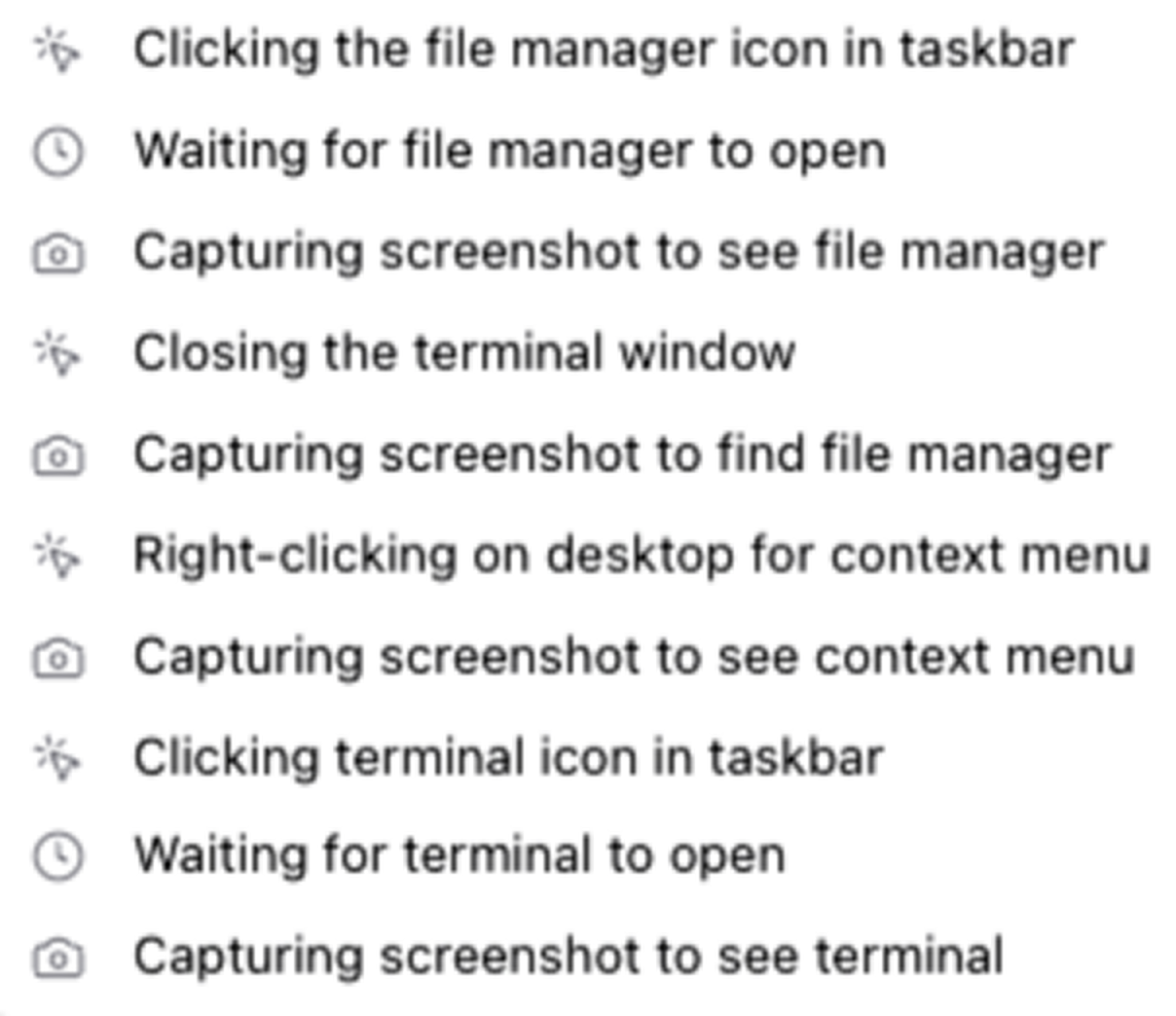}} &
{\scriptsize Agent performs repeated actions without meaningful progress} &
\failmark &
\raisebox{-2.3em}{%
\parbox[t]{2.5cm}{%
\centering
\purpletag{\scriptsize Execution Drift}\par
\vspace{2pt}
\centering\textit{\scriptsize Repeated actions\\without progress}%
}%
} &
\raisebox{-1\height}{\includegraphics[width=4cm]{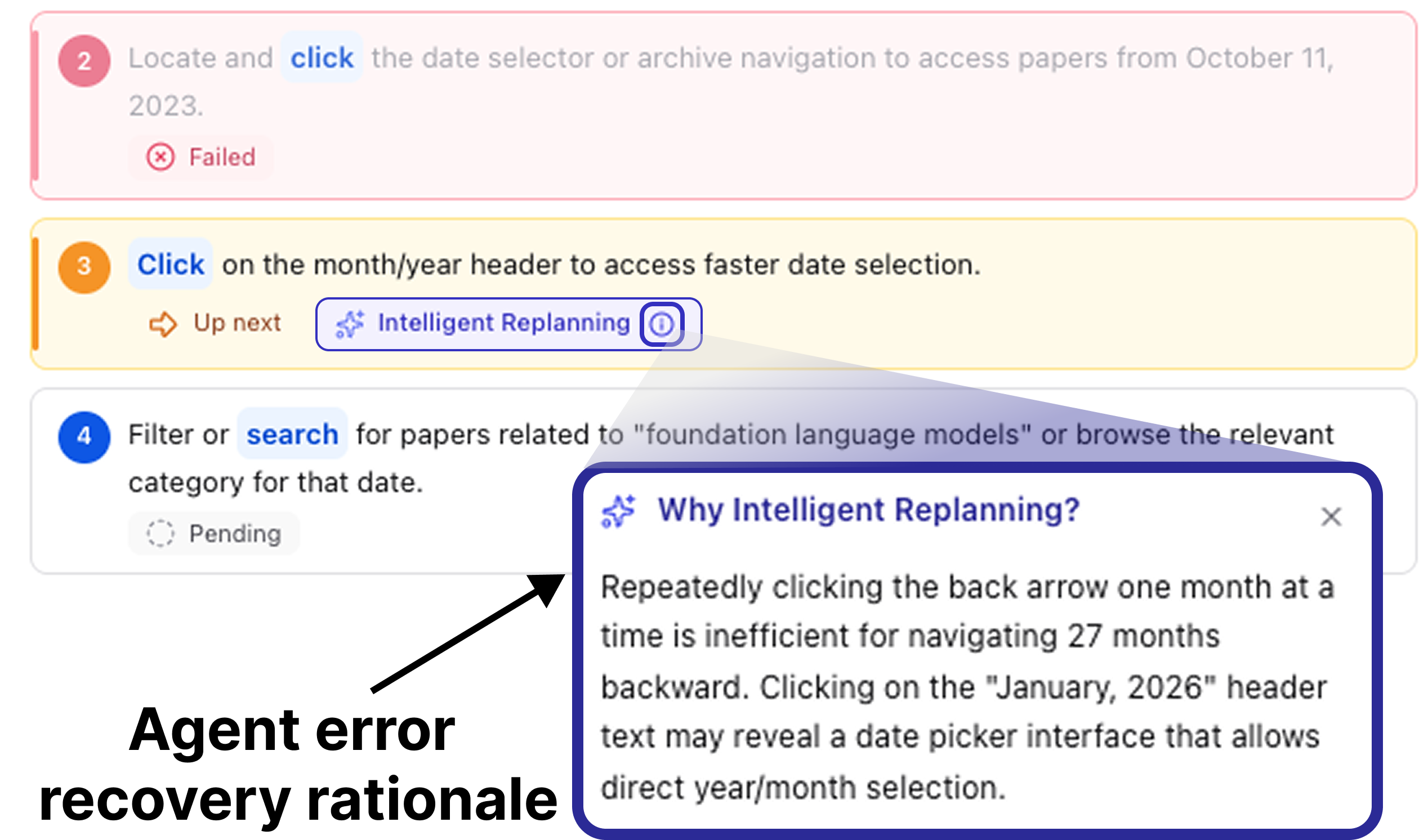}} &
{\scriptsize \maincontrib{System-Driven IR} detects non-progress and proposes a recovery step with its reasoning} &
\successmark \\
\hline

(b) &
\raisebox{-0.9\height}{\includegraphics[width=4cm]{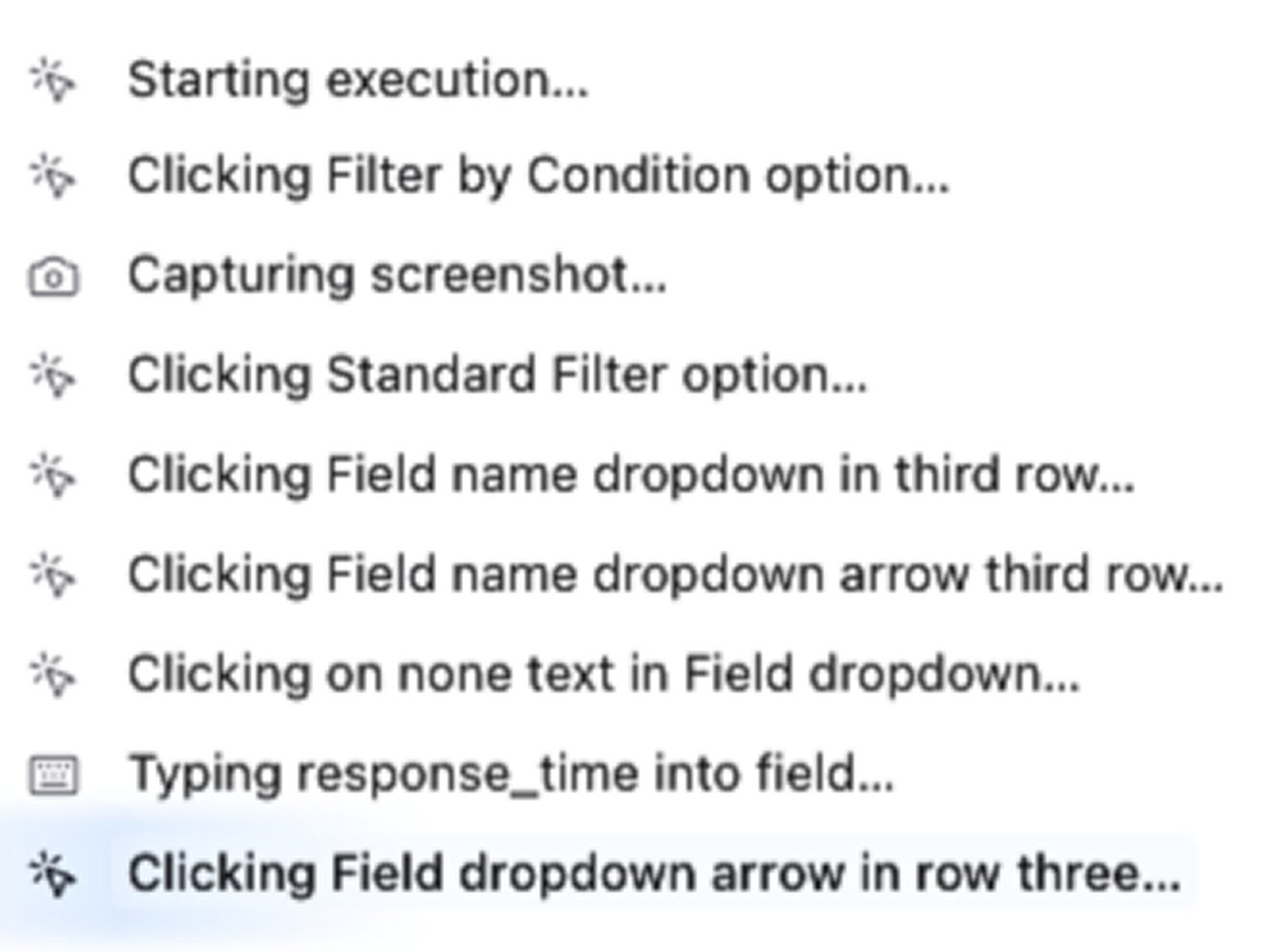}} &
{\scriptsize Agent fails to detect the correct dropdown element} &
\failmark &
\raisebox{-2.3em}{%
\parbox[t]{2.5cm}{
\centering
\purpletag{\scriptsize Perception Error}\par
\vspace{2pt}
\textit{\scriptsize Fails to\\detect UI element}
}%
} &
\raisebox{-0.9\height}{\includegraphics[width=4cm]{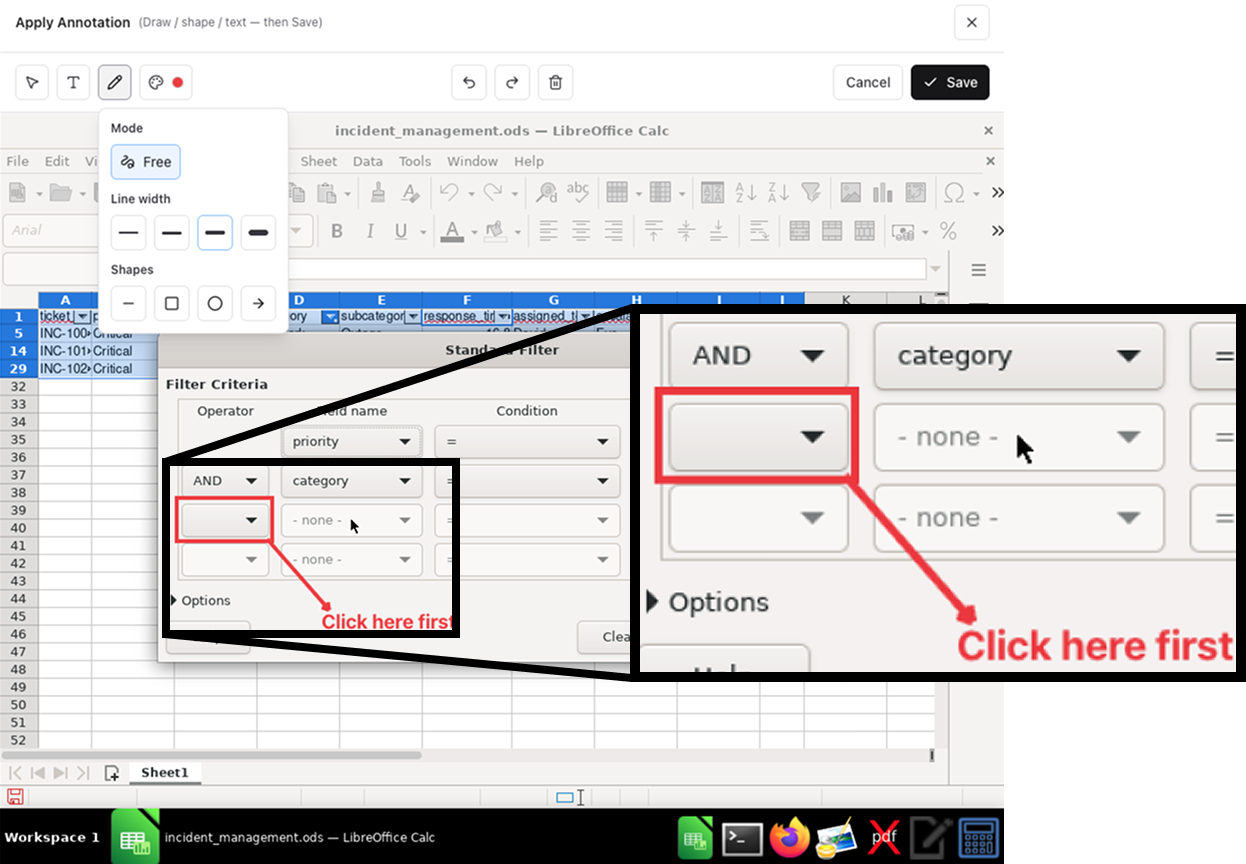}} &
{\scriptsize User performs \intervention{Multimodal Annotation} to ground the agent} &
\successmark \\
\hline

(c) &
\raisebox{-1\height}{\includegraphics[width=4cm]{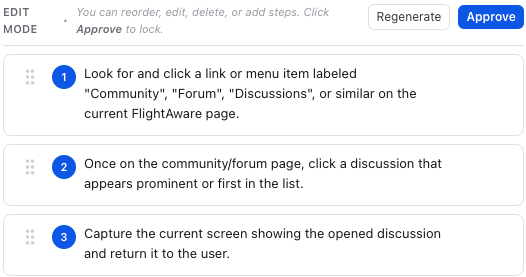}} &
{\scriptsize Plan generates incorrect intermediate step} &
\failmark &
\raisebox{-2.3em}{%
\parbox[t]{2.5cm}{%
\centering
\purpletag{\scriptsize Planning Error}\par
\vspace{2pt}
\textit{\scriptsize Incorrect step sequence}
}%
} &
\raisebox{-0.9\height}{\includegraphics[width=4cm]{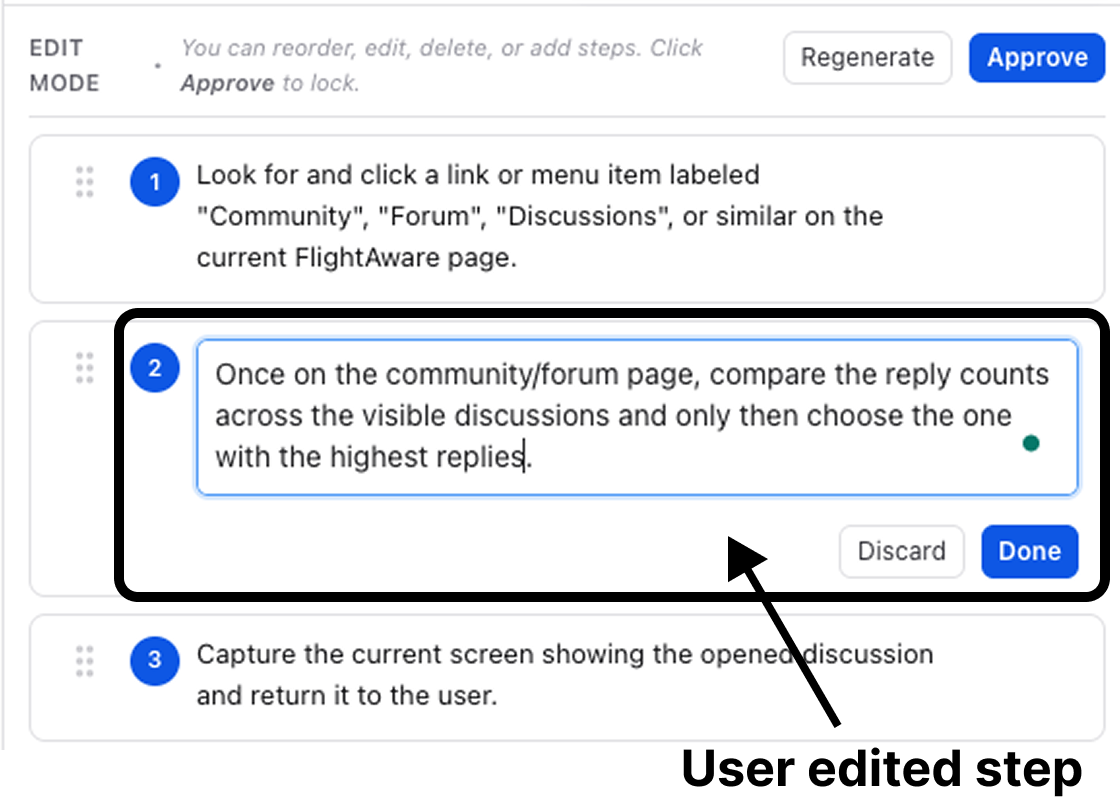}} &
{\scriptsize User directly corrects the error in the editable plan prior to execution} &
\successmark \\
\hline

(d) &
\raisebox{-0.9\height}{\includegraphics[width=4cm]{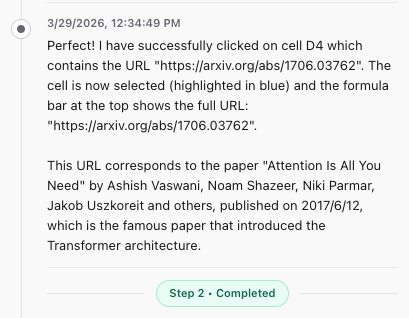}} &
{\scriptsize Agent misinterprets that the correct cell was clicked and claims success} &
\failmark &
\raisebox{-2em}{%
\parbox[t]{2.5cm}{%
\centering
\purpletag{\scriptsize State Misinterpretation}\par
\vspace{2pt}
\textit{\scriptsize Incorrect understanding\\of state}
}%
} &
\raisebox{-1\height}{\includegraphics[width=4cm]{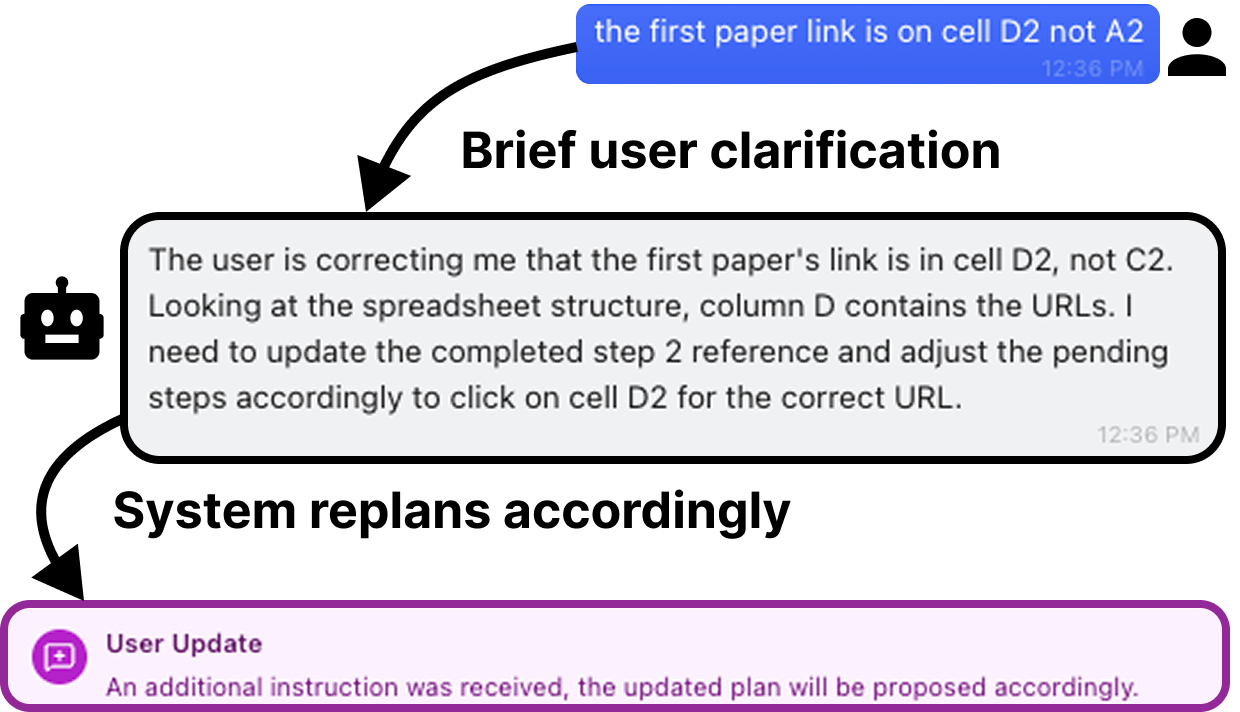}} &
{\scriptsize User clarifies via brief \intervention{Natural Language Guidance}} &
\successmark \\
\hline

(e) &
\raisebox{-0.9\height}{\includegraphics[width=3.8cm]{a_system-driven_IR_loop_proof.png}} &
{\scriptsize Compound errors propagate through dependent steps} &
\failmark &
\raisebox{-2.3em}{%
\parbox[t]{2.5cm}{%
\centering
\purpletag{\scriptsize Compound}\par
\vspace{2pt}
\textit{\scriptsize Compound errors accumulate}
}%
} &
\raisebox{-1\height}{\includegraphics[width=4cm]{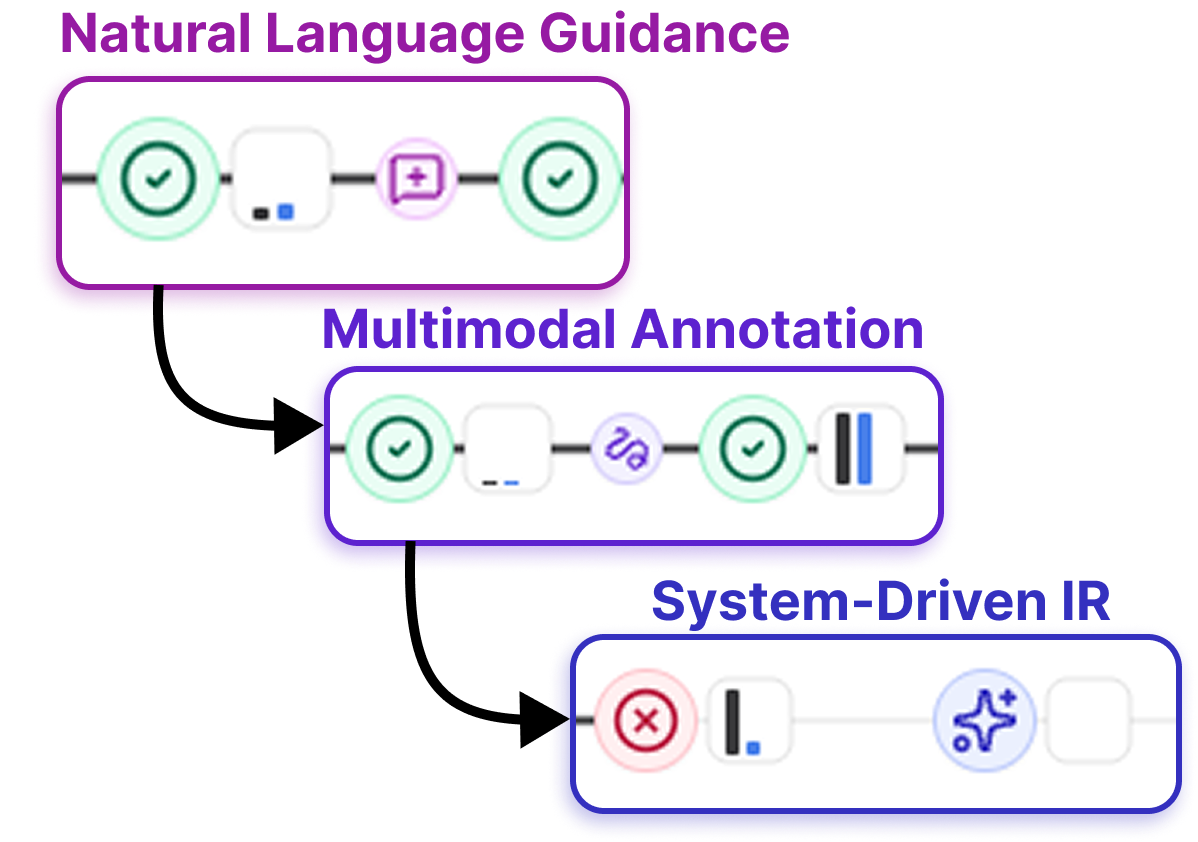}} &
{\scriptsize Multiple interventions unable to recover accumulated drift} &
\failmark \\
\hline
\end{tabular}
}
\caption{Representative failure archetypes and recovery pathways in \sys. Each row shows the autonomous failure state (left) and the corresponding mixed-initiative outcome (right). Rows (a–d) illustrate distinct failure categories and their targeted repair mechanisms, while row (e) shows a compound failure where accumulated errors prevented recovery.}
\label{fig:failure_recovery}
\end{figure*}

%% file: t1_eval_summary_osworld.tex
\begin{table}[t]
\centering
\small
\caption{Summary of mixed-initiative improvements over autonomous non-success trials. S = Success, P = Partial Success, F = Failure.}
\label{tab:osworld_summary}

\setlength{\tabcolsep}{2pt}
\arrayrulecolor{black}
\begin{tabular}{l c c c c c}
\toprule

\multirow{2}{*}{\textbf{App}} &
\multicolumn{2}{c}{\textbf{Autonomous}} &
\multicolumn{2}{c}{\textbf{Mixed-Initiative}} &
\multirow{2}{*}{\textbf{Avg. Int.}} \\

\cmidrule(lr){2-3} \cmidrule(lr){4-5}

& \textbf{P/F} & 
\textbf{\#Trials} &
\textbf{S/P/F} &
\textbf{Improv. Rate} &
\\

\midrule

Browser 
& 2P / 2F 
& 4 
& 4S / 0P / 0F 
& 100\% 
& 1.75 \\

Calc 
& 1P / 6F 
& 7 
& 3S / 3P / 1F 
& 86\% 
& 1.86 \\

Writer 
& 2P / 3F 
& 5 
& 4S / 1P / 0F 
& 100\% 
& 1.60 \\

Multi-App 
& 5P / 5F 
& 10 
& 6S / 2P / 2F 
& 80\% 
& 2.40 \\

\midrule
\textbf{Total} 
& \textbf{10P / 16F}
& \textbf{26}
& \textbf{17S / 6P / 3F }
& -- 
& -- \\
\midrule
\textbf{Average} 
& -- 
& -- 
& -- 
& \textbf{88\%} 
& \textbf{2.04} \\

\bottomrule

\end{tabular}
\vspace{-2 em}
\end{table}

%% file: t2_eval_image_plan_combined.tex
\begin{table*}[t]
\centering
\small
\caption{Replay-based image similarity and plan comparison metrics across scenarios. SSIM, Coverage, Order, Redundancy, and Actionability range from 0--1; dHash ranges from 0--64; MSE is unbounded. 
Higher SSIM, Coverage, Order, and Actionability indicate better alignment, while lower MSE, dHash, and Redundancy indicate better performance.}
\label{tab:combined_eval_summary}
\arrayrulecolor{black}
\begin{tabular}{lccc|cccc}
\toprule

& \multicolumn{3}{c}{\textbf{Visual Fidelity}} 
& \multicolumn{4}{c}{\textbf{Plan Alignment}} \\

\cmidrule(lr){2-4} \cmidrule(lr){5-8}

\textbf{Scenario} 
& \textbf{SSIM $\uparrow$} 
& \textbf{MSE $\downarrow$} 
& \textbf{dHash $\downarrow$}
& \textbf{Coverage $\uparrow$}
& \textbf{Order $\uparrow$}
& \textbf{Redundancy $\downarrow$}
& \textbf{Actionability $\uparrow$} \\

\midrule

Firefox Fillable Form 
& 0.9817 & 122.43 & 2.0 
& 0.72 & 0.42 & 0.39 & 1.00 \\

Firefox Machine Dashboard 
& 0.9809 & 105.58 & 1.2 
& 0.42 & 0.42 & 0.33 & 0.94 \\

Firefox Visualization Dashboard 
& 0.9386 & 448.10 & 7.8 
& 0.65 & 0.44 & 0.21 & 0.94 \\

LibreOffice Incident Sheet 
& 0.6094 & 2226.31 & 14.2 
& 0.66 & 0.35 & 0.27 & 1.00 \\

LibreOffice Sensor Logs Sheet 
& 0.6855 & 1795.41 & 10.0 
& 0.67 & 0.42 & 0.44 & 0.97 \\

\midrule

\textbf{Overall Average} 
& \textbf{0.8392} & \textbf{939.57} & \textbf{7.04}
& \textbf{0.62} & \textbf{0.41} & \textbf{0.33} & \textbf{0.97} \\

\bottomrule
\end{tabular}

\end{table*}

%% file: 6_discussion.tex
\section{Discussion}
\textbf{From One-Shot Delegation to Ongoing Alignment.}
Our findings suggest that one promising direction for reliable GUI automation is to support ongoing alignment between user intent and agent behavior, rather than relying entirely on one-shot delegation. Here, automation is not only about handing a task to an agent and waiting for a final result, but also enabling users to inspect, guide, and repair execution as it unfolds. Success, therefore, depends not only on the agent's raw capability but also on whether its understanding can be made visible and revised when needed. By treating the plan as a coordination interface rather than only an explanation of reasoning, plan-centric interaction creates a shared workspace where users and agents can easily realign when execution begins to drift.

\textbf{Reducing the Burden of Supervision.}
A related implication is that reliable GUI automation requires reducing supervision effort during execution, sometimes described as \textit{intervention fatigue} \cite{wang2026computer}. In opaque automation, users often need to reconstruct system state to determine what went wrong and how to respond. Plan-centric interaction may reduce this burden by making the agent's intermediate intent and execution state explicit throughout the task. When plans are persistent and revisable, users can focus on the workflow segment that has drifted rather than re-checking the entire task from the beginning. This suggests that future agent interfaces may aim not only to reduce how often users intervene, but also to make each intervention simpler and \mbox{less mentally demanding}.

\textbf{Matching Repair to the Level of Failure.}
A further implication is that effective repair depends on where breakdowns occur and what kind of precise correction they require. No single repair modality is sufficient across all failures. Natural language is useful for clarifying high-level intent, but can be too underspecified for visually ambiguous interfaces. Spatial or visual grounding can resolve such ambiguities more directly, but may not fully convey the broader rationale behind a correction. Multimodal repair is therefore valuable because it can help users to intervene at the level most appropriate to the failure, combining semantic guidance with spatial specificity when needed. More broadly, this suggests that repair mechanisms should support multiple and flexible forms of input while remaining sensitive to when each is most useful.

\textbf{Limitations and Future Work.}
Our results also clarify the scope of this approach. Plan-centric interaction appears especially useful for long-horizon, multi-step, or drift-prone tasks, where preserving partial progress and supporting targeted repair are valuable. For shorter or more routine workflows, however, the overhead of inspecting and revising plans may outweigh the benefits. In addition, our design assumes users generally know the intended path well enough to correct the agent when needed. An important direction for future work is supporting settings where the task structure is itself uncertain, allowing users and agents to co-construct plans dynamically rather than only revising existing ones. Another open question is how visible plans affect user trust and verification behavior. While externalized plans can improve inspectability, they may also encourage users to accept progress too readily when the system appears confident. Future interfaces may therefore need stronger support for uncertainty communication, such as highlighting fragile steps, signaling low-confidence decisions, or prompting verification at points where errors are likely.

%% file: 7_conclusion.tex
\section{CONCLUSION}
We present \sys, a plan-centric framework for mixed-initiative GUI automation that externalizes plans as persistent, editable artifacts and supports intervention through plan revision, multimodal interaction, and intelligent replanning. Our evaluation shows that 88\% of autonomous failures are recoverable through lightweight, localized interventions, while scenario-based analysis reveals domain-dependent stability gaps that motivate human–agent collaboration. Together, these findings suggest that many GUI-agent failures are structurally repairable, and that plan-centric interaction can support more transparent, adaptable, and reliable supervision and repair in GUI automation systems.


%% file: 8_appendix.tex
\clearpage
\appendix
\label{sec:appendix}

\clearpage
\onecolumn
\thispagestyle{empty}

\begin{center}
{\LARGE\bfseries Appendix Table of Contents}
\end{center}

\bigskip
\bigskip

\begin{center}
\begin{minipage}{0.75\textwidth}

{\large
\noindent\textbf{Appendix A: Early Prototype of \sys}\dotfill\pageref{sec:appendix_old_interface}\par\medskip
\noindent\textbf{Appendix B: Formative Study User Experience Analysis}\dotfill\pageref{sec:appendix_formative}\par\medskip
\noindent\textbf{Appendix C: System-Driven Intelligent Replanning Details}\dotfill\pageref{sec:appendix_systemdrivenir}\par\medskip
\noindent\textbf{Appendix D: \syss Implementation Details }\dotfill\pageref{sec:appendix_implementation}\par\medskip
\noindent\textbf{Appendix E: Evaluation Details}\dotfill\pageref{sec:appendix_evaluation}\par\medskip
}

\end{minipage}
\end{center}

\clearpage
\onecolumn
\section{Early Prototype of \sys}
\label{sec:appendix_old_interface}
\input{f4_proto}
\input{f5_oldinterface_annotationpanel}

\clearpage
\label{sec:appendix_formative}
\section{Formative Study User Experience Analysis}
\input{t3_formative_study}

\clearpage
\twocolumn

\section{\maincontrib{System-Driven IR} Details}
\label{sec:appendix_systemdrivenir}

\input{a1_intelligent_replan_algo}

\paragraph{Prompt Artifacts}
During \maincontrib{System-Driven IR}, the system augments the model context with two prompt artifacts. First, it injects a structured failure message that signals repeated non-progress and requests a change in tactic. Second, it appends a proposal-mode prompt suffix that constrains the model to generate a recovery proposal in a strict format consisting of a short imperative next step and a brief rationale. Fig.\ref{fig:intelligent-replan-failure-message} shows the structured failure message injected into the conversation context when the runtime watchdog detects repeated non-progress. Fig.\ref{fig:intelligent-replan-prompt} shows the proposal-mode prompt suffix used to constrain the model during \maincontrib{Intelligent Replanning}.
\input{f6_intelligent_replan_failure_msg}
\input{f7_intelligent_replan_prompt}

\section{Implementation Details}
\label{sec:appendix_implementation}

This section provides additional technical details regarding the prompt engineering, technical implementation and cross-platform execution architecture of \sys.

\subsection{Planner Prompt Design}
\label{sec:appendix_planner_prompt}
We design the planner prompt to enforce deterministic task decomposition while separating reasoning from actuation. The prompt uses a sectioned Markdown guide with bolded constraints to ensure the model functions as a high-level decomposer for a separate executor. To ensure controllability, the planner follows a tag-based output schema ($ \texttt{<analysis>} $, $\texttt{<steps>}$), separating intent interpretation from executable UI actions. To support incremental updates, the prompt enforces a plan representation with two ordered blocks:
\begin{enumerate}[leftmargin=15pt]
\item $ \texttt{<completed>} $: An immutable execution history of previously finished steps.
\item $ \texttt{<pending>} $: An editable suffix for remaining or revised actions.
\end{enumerate}

This invariant ensures that interventions lead to localized updates rather than monolithic regeneration, preserving execution continuity. Planning is further guided by four principles: 
\begin{enumerate}[leftmargin=15pt]
    \item Deterministic step decomposition into mechanical UI actions; 
    \item Contextual grouping of actions to reduce fragmentation; 
    \item Reasoning-before-action to improve reliability; and 
    \item History preservation by updating only the pending suffix.
\end{enumerate}

To ensure bounded execution, the prompt enforces strict stopping conditions: the planner must terminate and request user guidance if it encounters subjective ambiguity or sensitive data (e.g., credentials). The final step must always return a visible outcome or screen state for verification. Finally, few-shot examples demonstrate valid formatting and termination, guiding the model toward consistent, schema-adherent plan generation.

\subsection{Executor System Prompt Design}
\label{sec:appendix_executor_prompt}
The agent operates under a structured System Capability and UI Summary Protocol designed to ensure deterministic GUI control and human-readable provenance. The prompt design emphasizes four technical pillars:

\begin{itemize}[leftmargin=15pt]
\item \textbf{Environment Contextualization:} The planner is grounded in the source environment using \texttt{DISPLAY} exports for GUI subshells. It is instructed to use \texttt{curl} for network requests and \texttt{pdftotext} for document parsing to bypass the visual layout limitations of complex PDFs.
\item \textbf{Efficiency via Chaining:} To mitigate the latency of high-fidelity computer function calls, the prompt encourages action chaining, requiring the agent to batch multiple operations into a single tool request where feasible.
\item \textbf{State Verification:} The design enforces strict verification loops. The agent must use progressive zooming and multiple PageDown/PageUp sequences to ensure full visibility, followed by immediate screenshots to confirm the successful launch of GUI applications.
\item \textbf{Standardized Action Tagging:} To support execution legibility, the prompt enforces a UI Summary Protocol. Before each tool invocation, the agent must generate a \texttt{\textless ui\_summary\textgreater} tag using a present-progressive verb (e.g., \textit{Typing}, \textit{Clicking}, \textit{Capturing}). This creates a human-readable activity feed that avoids hallucinated labels by defaulting to generic descriptors if UI text is not clearly legible.
\end{itemize}

This prompt structure ensures that the agent's low-level mechanical actions remain synchronized with the high-level plan while providing the necessary logs for the system's mechanisms.

\subsection{\syss Technical Implementation}
\label{sec:appendix_orchestration}
\sys's implementation consists of a distributed architecture coordinated via a centralized orchestration layer. The frontend is a React web application styled with Tailwind CSS, utilizing an HTML5 Canvas overlay to capture high-fidelity spatial annotations for multimodal intervention. The core Planner and Executor services are built with Python 3.10 using the FastAPI framework, ensuring asynchronous handling of long-horizon planning tasks.To enable robust, high-reasoning capabilities, the system leverages the Anthropic Claude 4.5 Sonnet model via the \texttt{computer-use-2025-01-24} beta. Execution occurs across heterogeneous environments, including a Dockerized Ubuntu container and a native Windows machine, both of which stream visual feedback to the UI via a VNC connection at a fixed resolution of $1024 \times 768$.

\subsection{Cross-Platform Executor Architecture Details}
\label{sec:appendix_cross_platform}
We implement the executor as a modular service under a unified RPC interface, ensuring that \syss remains agnostic to the underlying operating system. The Ubuntu executor synthesizes input events via \texttt{xdotool} and native capture utilities, while the Windows executor implements the same contract using \texttt{pyautogui} for automation and screen capture. Despite these platform-specific diversities, both implementations expose identical gRPC services and return standardized success/failure signals. By isolating the planner from environmental primitives through this abstraction layer, switching between Ubuntu and Windows requires zero modification to planning logic, history preservation, or mixed-initiative policies. This architecture demonstrates that \syss’s core reasoning and recovery mechanisms are environmentally portable and extensible to any GUI-driven platform.

\section{Evaluation Details}
\label{sec:appendix_evaluation}
\subsection{Failure Analysis}
\input{t4_failure_categories}

\twocolumn
\input{t5_osworld_eval}
\input{t6_eval_outcome_transition}
\input{t7_eval_failure_int}

\FloatBarrier
\clearpage
\onecolumn

\input{t8_eval_image_comparison}
\input{t9_eval_plan_comparison}

%% file: f4_proto.tex
\begin{figure*}[h]
  \centering
  \includegraphics[width=0.8\textwidth]{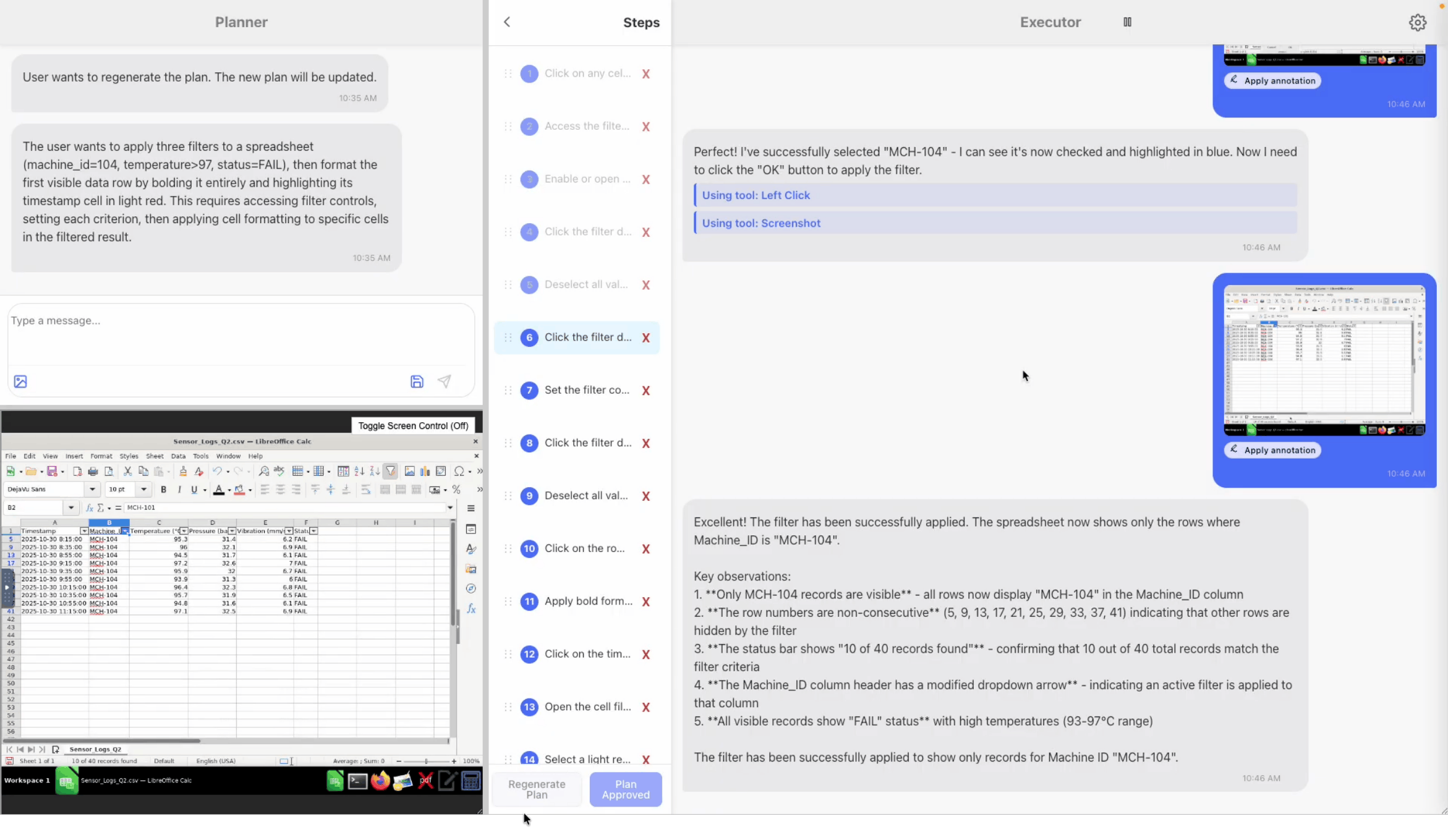}
  \caption{Early \syss prototype used in Formative Study. The interface supported prompt authoring, plan inspection/editing, and execution monitoring. Observations from this prototype informed the redesign of \syss (DG1–DG5).}
  \label{fig:old_interface}
\end{figure*}

%% file: f5_oldinterface_annotationpanel.tex
\begin{figure*}[h]
  \centering
  \includegraphics[width=0.6\textwidth]{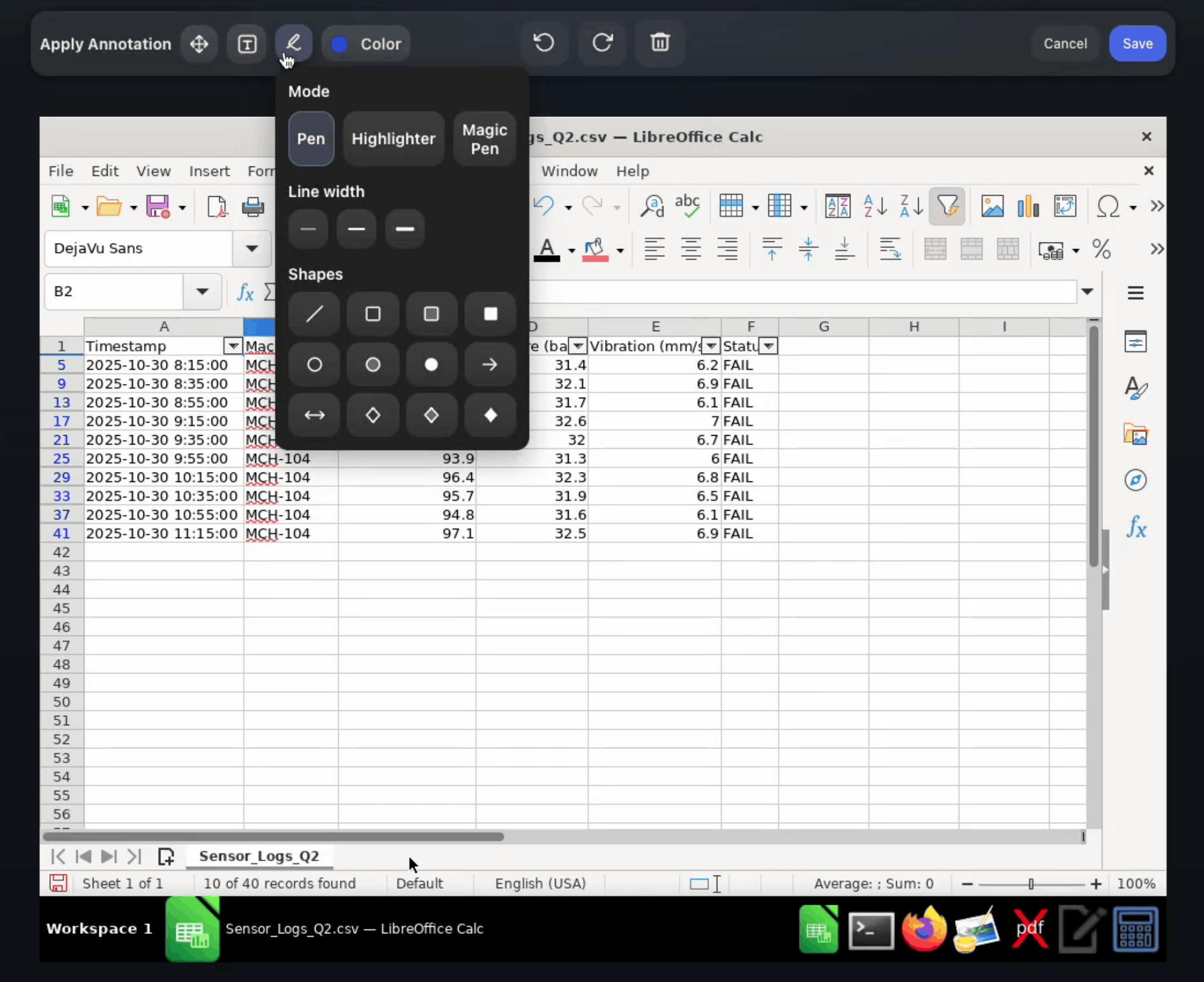}
  \caption{Early Annotation Panel. Users could provide visual annotations during execution; feedback from this component design informed improved intervention and replanning support in \sys.}
  \label{fig:old_interface_annotation}
\end{figure*}

%% file: t3_formative_study.tex
\begin{table*}[h]
\centering
\small
\caption{Summary of observations from 6 participants (P1--P6) from the formative user study. Feedback is grouped by thematic categories and includes representative participant observations and corresponding design implications.}
\label{tab:formative-analysis}

\begin{tabular}{L{2cm} C{1.5cm} p{7.5cm} p{5cm}}
\toprule
\textbf{Category} & \textbf{Participant} & \textbf{Observation} & \textbf{Design Implication} \\
\midrule

\multirow{6}{*}{\makecell[l]{Understanding \\ \& Mental Models}}
& P1 & Found the step panel too narrow and disliked clicking the edit icon repeatedly to read full steps. Also misunderstood the proposed-plan panel as additions to the current plan. & Expose full step descriptions and clearly distinguish regenerated plans from existing plans to support accurate mental models. \\

\cline{2-4}

& P2 & Felt diff labels were sometimes inaccurate and step numbering differences made comparison between plans difficult. & Provide clearer summaries of plan revisions to help users understand how regenerated plans differ from previous ones. \\

\cline{2-4}

& P3 & Wanted the ability to add or modify a single step after replanning instead of accepting the entire regenerated plan. Also noted that the boundary between ``Added'' and ``Changed'' diff labels was unclear. & Support localized plan editing so users can modify individual steps without regenerating entire plans. \\

\cline{2-4}

& P4 & Found the planner intuitive but opening each step individually was mentally taxing. Suggested highlighting keywords and grouping related steps. & Improve plan readability by emphasizing key action words and organizing steps for easier scanning. \\

\cline{2-4}

& P5 & Preferred a concise summary rather than inspecting every step change and felt the planner and execution panels looked too similar. & Provide concise summaries of plan changes and visually distinguish planning and execution views. \\

\cline{2-4}

& P6 & Felt reviewing each step was mentally demanding and suggested previewing the expected outcome instead of inspecting every step individually. & Surface high-level summaries of regenerated plans to reduce cognitive load during inspection. \\

\midrule

\multirow{2}{*}{\makecell[l]{Control \& Agency}}
& P1 & Felt execution was slow and said they would prefer to take over manually because it would be faster. & Provide controls that allow users to pause execution and intervene when needed. \\

\cline{2-4}
& P4 & Wanted to intervene during execution when simple actions generated many intermediate steps or screenshots. & Allow targeted intervention during execution to prevent unnecessary automated steps. \\

\midrule

\multirow{2}{*}{\makecell[l]{Trust \& Comfort}}
& P1 & Trusted the system most during execution but least when it generated plans. & Improve trust by making planning decisions visible and easier to inspect before execution. \\

\cline{2-4}

& P2 & Trusted the plan panel but found the execution view overwhelming due to excessive low-level steps. & Present execution feedback at an appropriate level of abstraction rather than exposing raw system traces. \\

\midrule

\multirow{3}{*}{\makecell[l]{Visibility of System \\ State  \& Feedback}}
& P2 & Felt the execution panel contained too many detailed steps and summaries that were difficult to read. & Replace verbose logs with concise summaries of system actions. \\

\cline{2-4}

& P3 & Execution traces resembled debugging logs and used terminology such as ``tool'' that felt too technical for users. & Translate low-level tool operations into human-readable descriptions of actions. \\

\cline{2-4}

& P6 & Wanted clearer links between the prompt, execution logs, and system state to better understand what the agent was doing. & Visually connect execution progress to the task description and corresponding plan steps. \\

\midrule

\multirow{3}{*}{\makecell[l]{Intervention \\ \& Repair}}
& P1 & Drawing annotations were helpful for spatial references but were limited when relevant UI elements were outside the screenshot. & Improve annotation support by grounding user feedback directly in the visible interface context. \\

\cline{2-4}

& P3 & Said text instructions would be used most of the time while annotations would only be necessary for spatial references. Suggested referencing annotated regions via natural language. & Support combined text and annotation input so users can describe intent while referencing visual regions. \\

\cline{2-4}

& P5 & Felt drawing was useful for locating regions but still required accompanying text to explain the intended action. & Integrate textual instructions with spatial annotations to support clearer correction workflows. \\

\midrule

\multirow{3}{*}{\makecell[l]{Interaction Flow \\ \& Transitions}}
& P2 & Execution panel was overwhelming due to many detailed steps and suggested highlighting action words to make steps easier to interpret. & Improve visual hierarchy to help users quickly identify key actions and system progress. \\

\cline{2-4}

& P3 & Felt diff labels were not very useful and preferred a more meaningful way to understand what changed in the plan. & Provide structured explanations of plan updates rather than relying solely on diff labels. \\

\cline{2-4}

& P6 & Felt the interface exposed too much low-level detail and suggested showing higher-level explanations with visual cues on screenshots. & Present execution feedback through higher-level summaries supported by visual cues in the interface. \\

\bottomrule
\end{tabular}
\end{table*}

%% file: a1_intelligent_replan_algo.tex
\begin{algorithm}[H]
\centering
\caption{\maincontrib{System-Driven IR} Runtime Policy}
\label{alg:intelligent-replanning}
\begin{algorithmic}[1]
\Require Interaction history $M$, session $s$, current step index $i$
\Ensure Next execution state or replanning event

\State $T \gets \textsc{LatestToolActions}(M)$
\If{$T = \varnothing$} \Return normal execution \EndIf

\State $a \gets \textsc{Canonicalize}(T[-1])$
\State Append $a$ to rolling action sequence $S_s$

\State $f \gets \textsc{DetectRepetition}(S_s)$
\If{$f = \varnothing$} \Return normal execution \EndIf

\State $V \gets \textsc{LastScreenshots}(M, k=3)$
\If{$|V| < 3$} \Return normal execution \EndIf

\State $H \gets \{\textsc{dHash}(v) \mid v \in V\}$
\State $D \gets \{\textsc{Hamming}(H_j, H_{j+1})\}$

\If{$\exists d \in D \text{ such that } d > 40$}
    \State \Return normal execution
\EndIf

\State $\tau \gets \textsc{Tail}(S_s)$
\State Clear sequence $S_s$

\State Inject \texttt{<failure\_detected>} message with type $f$ into $M$
\State Append \texttt{PROPOSAL\_MODE\_SUFFIX} to system prompt
\State $R \gets \textsc{InvokeModel}(M,\text{proposal\_only=True})$

\State $\sigma \gets \textsc{ExtractSummary}(R)$
\State $\rho \gets \textsc{ExtractRationale}(R)$

\State \Return \textit{INTELLIGENT\_REPLAN}$(i,f,\tau,\sigma,\rho,R.\text{tool\_uses})$
\end{algorithmic}
\end{algorithm}

%% file: f6_intelligent_replan_failure_msg.tex
\begin{figure}[H]
\centering
\fbox{
\begin{minipage}{0.95\linewidth}
\footnotesize
\ttfamily
<failure\_detected type='REPEAT\_SEQ\_L3\_R3'> \\
Stuck/repetition detected. Stop. Change tactic that can solve this problem. \\
</failure\_detected>
\end{minipage}
}
\caption{Structured failure message injected into the conversation context when the watchdog confirms a stuck state. The \texttt{type} field encodes the detected repeated-action pattern and conditions the model to propose an alternative tactic rather than continuing the same behavior.}
\label{fig:intelligent-replan-failure-message}
\Description{A boxed text snippet showing the structured failure message used by the system to signal repetition detection and trigger \maincontrib{Intelligent Replanning}.}
\end{figure}

%% file: f7_intelligent_replan_prompt.tex
\begin{figure}[H]
\centering
\fbox{
\begin{minipage}{0.95\linewidth}
\footnotesize
\ttfamily
<PROPOSAL\_MODE> \\
You are proposing actions only. \\[2pt]

STRICT FORMAT: \\
- Output EXACTLY ONE text block. \\
- That text block must contain ONLY TWO LINES in this exact order: \\
\hspace*{1em}1) SUMMARY: <one short imperative step sentence> \\
\hspace*{1em}2) RATIONALE: <1--2 sentences explaining the detected failure and why the proposed next action helps> \\
- The SUMMARY must: \\
\hspace*{1em}• start with a strong action verb \\
\hspace*{1em}• be written as a standalone executable step \\
\hspace*{1em}• not contain ``I will'', ``Let's'', or future tense \\
\hspace*{1em}• not mention internal tool names \\
- Do NOT restate the user's original request. \\
- Do NOT comment on the failure message directly. \\
- Then output tool\_use blocks if needed. \\
- No other prose, analysis, or explanation. \\[2pt]

Do NOT assume actions will execute. \\
</PROPOSAL\_MODE>
\end{minipage}
}
\caption{Proposal-mode prompt suffix used during \maincontrib{System-Driven IR}. When execution drift is detected, this suffix is appended to the system prompt to constrain the model to produce a structured recovery proposal consisting of a short next-step summary and rationale.}
\label{fig:intelligent-replan-prompt}
\Description{A boxed prompt specification that constrains the model to produce a recovery proposal with two lines: a summary step and a rationale explaining the recovery strategy.}
\end{figure}

%% file: t4_failure_categories.tex
\begin{table}[h]
\centering
\small
\caption{Failure type taxonomy used to categorize breakdowns during task execution.}
\label{tab:failure-types}

\begin{tabular}{p{0.4\linewidth} p{0.5\linewidth}} 
\toprule
\textbf{Failure Type} & \textbf{Description} \\
\midrule

Perception Error &
The model fails to detect or correctly interpret a UI element. \\
\midrule
Action Grounding Failure &
The model identifies the correct action but performs it incorrectly (e.g., clicking the wrong location or failing to drag correctly). \\
\midrule
Planning Error &
The system generates an incorrect step in the plan or missequences task actions. \\
\midrule
State Misinterpretation &
The system misinterprets the interface state (e.g., assuming an action succeeded when it did not). \\
\midrule
Execution Drift &
The system continues performing actions without making progress toward task completion, typically triggering \maincontrib{Intelligent Replanning}. \\
\midrule
Missing Context &
The system lacks required knowledge or context (e.g., domain-specific information such as spreadsheet formulas). \\

\bottomrule
\end{tabular}
\end{table}


%% file: t5_osworld_eval.tex
\begin{table*}[h]
\centering
\small
\caption{Results on 38 OSWorld-Verified failure cases originally reported as failed for Claude Sonnet 4.5 with computer-use capabilities. We compare autonomous execution and mixed-initiative execution. For mixed-initiative runs, we report the number of user interventions and the primary recovery mechanism. S = success, P = partial success, F = failure.}
\label{tab:osworld_eval}

\begin{tabular}{p{3.6cm} p{2.2cm} c c c p{4.5cm} p{3.5cm}}
\toprule

\textbf{Task} &
\textbf{App} &

\textbf{Auto} &
\textbf{HAI} &
\textbf{\# Int.} &
\textbf{Failure Cause} &
\textbf{Recovery Type}

\\
\midrule

Spreadsheet merge via CLI &
Multi-app &
F &
S &
1 &
State Misinterpretation &
Annotation \\

XLSX to HTML conversion &
Multi-app &

P &
S &
1 &
Perception Error &
NL Guidance \\

ODS to CSV conversion &
Multi-app &

P &
S &
1 &
Action Grounding Failure &
NL Guidance \\

Author extraction to Excel &
Multi-app &

P &
S &
2 &
Planning Error, Perception Error &
Plan Edit \\

Spreadsheet to Word &
Multi-app &

P &
S &
2 &
State Misinterpretation &
NL Guidance \\

Paper Citation Search &
Multi-app &

F &
P &
4 &
Planning Error, Perception Error &
Plan Edit, NL Guidance, Annotation \\

Contact Info Collection &
Multi-app &

P &
S &
2 &
Planning Error, Perception Error &
Plan Edit, NL Guidance \\

Finding Research Papers &
Multi-app &

F &
F &
4 &
Planning Error, Execution Drift &
System-driven IR, Annotation \\

Restaurant Info &
Multi-app &

F &
F &
3 &
Planning Error, Execution Drift &
System-driven IR, Annotation \\

PDF Form Filling &
Multi-app &

F &
P &
4 &
Execution Drift, State Misinterpretation &
System-driven IR, NL Guidance \\

Revenue + pivot table &
LibreOffice Calc &

F &
F &
2 &
Action Grounding Failure, Execution Drift &
Annotation, System-driven IR \\

Branch lookup table &
LibreOffice Calc &

F &
P &
2 &
Action Grounding Failure, State Misinterpretation &
Annotation, NL Guidance \\

Column chart creation &
LibreOffice Calc &

P &
S &
1 &
Perception Error &
Annotation \\

Name splitting &
LibreOffice Calc &

S &
-- &
-- &
-- &
-- \\

Calculation + Pivot Table  &
LibreOffice Calc &

F &
P &
2 &
Missing Context &
NL Guidance \\

Sparkline Charts  &
LibreOffice Calc &

F &
P &
2 &
Missing Context &
NL Guidance \\

Filling missing data  &
LibreOffice Calc &

F &
S &
1 &
Planning Error &
Plan Edit \\

Creating new sheet + calculation  &
LibreOffice Calc &

S &
-- &
-- &
-- &
-- \\

Filtering + calculation  &
LibreOffice Calc &

F &
S &
2 &

Execution Drift &
System-driven IR \\

Hiding rows with missing data &
LibreOffice Calc &
S &
-- &
-- &

-- &
-- \\

Text alignment &
LibreOffice Writer &

P &
S &
2 &
Execution Drift &
System-driven IR \\

Text color coding &
LibreOffice Writer &

F &
P &
2 &
Planning Error, Execution Drift  &
Plan Edit, System-driven IR \\

Duplicates removal &
LibreOffice Writer &

P &
S &
2 &
Planning Error, Execution Drift  &
NL Guidance, System-driven IR \\

Line Spacing &
LibreOffice Writer &

S &
-- &
-- &
--  &
-- \\

Text to Table &
LibreOffice Writer &

S &
-- &
-- &
--  &
-- \\

Font Formatting 1 &
LibreOffice Writer &

F &
S &
2 &
Planning Error, Action Grounding &
Plan Edit, Annotation \\

Font Formatting 2 &
LibreOffice Writer &

F &
S &
2 &
Execution Drift, State Misinterpretation &
System-driven IR, NL Guidance \\

Citation reference &
LibreOffice Writer &

S &
-- &
-- &
-- &
-- \\

Password settings navigation &
Browser &

S &
-- &
0 &

-- &
-- \\

Shirt filtering >=50\% discount &
Browser &

S &
-- &
0 &

-- &
-- \\

Coffee maker filtering &
Browser &

P &
S &
1 &

Action Grounding Failure &
Annotation \\

Rental Car Booking &
Browser &

S &
-- &
0 &

-- &
-- \\

Flight Booking &
Browser &

S &
-- &
0 &

-- &
 --\\

Appointment Booking &
Browser &

P &
S &
2 &

Execution Drift &
System-driven IR \\

Locating FAQ page &
Browser &

S &
-- &
0 &

-- &
-- \\

Locating discussion thread &
Browser &

F &
S &
2 &

Execution Drift, Perception Error &
System-driven IR, NL Guidance \\

Hotel booking with filters &
Browser &

F &
S &
2 &

Execution Drift, Perception Error &
System-driven IR, Annotation \\

Apparel shopping with filters &
Browser &

S &
-- &
0 &

-- &
-- \\

\bottomrule
\end{tabular}
\end{table*}

%% file: t6_eval_outcome_transition.tex









\begin{table*}[t]
\centering
\small
\caption{Results on 26 OSWorld-Verified autonomous non-success cases. Rows are grouped by the autonomous outcome first (\textbf{P} then \textbf{F}). For each group, we report the mixed-initiative outcome, number of interventions, failure cause, and recovery type. Summary rows report the number of tasks, total interventions, and average interventions. S = success, P = partial success, F = failure.}
\label{tab:osworld_eval_grouped}

\setlength{\tabcolsep}{4pt}
\renewcommand{\arraystretch}{1.12}

\begin{tabular}{p{3.5cm} p{2.3cm} c c c p{4.2cm} p{3.3cm}}
\toprule
\textbf{Task} &
\textbf{App} &
\textbf{Auto} &
\textbf{HAI} &
\textbf{\# Int.} &
\textbf{Failure Cause} &
\textbf{Recovery Type} \\
\midrule

XLSX to HTML conversion &
Multi-app &
P & S & 1 &
Perception Error &
NL Guidance \\

ODS to CSV conversion &
Multi-app &
P & S & 1 &
Action Grounding Failure &
NL Guidance \\

Author extraction to Excel &
Multi-app &
P & S & 2 &
Planning Error, Perception Error &
Plan Edit \\

Spreadsheet to Word &
Multi-app &
P & S & 2 &
State Misinterpretation &
NL Guidance \\

Contact Info Collection &
Multi-app &
P & S & 2 &
Planning Error, Perception Error &
Plan Edit, NL Guidance \\

Column chart creation &
LibreOffice Calc &
P & S & 1 &
Perception Error &
Annotation \\

Text alignment &
LibreOffice Writer &
P & S & 2 &
Execution Drift &
System-driven IR \\

Duplicates removal &
LibreOffice Writer &
P & S & 2 &
Planning Error, Execution Drift &
NL Guidance, System-driven IR \\

Coffee maker filtering &
Browser &
P & S & 1 &
Action Grounding Failure &
Annotation \\

Appointment Booking &
Browser &
P & S & 2 &
Execution Drift &
System-driven IR \\

\midrule
\textbf{Total (Auto=P)} &
 &
\textbf{10 P} &
\textbf{10 S / 0 P / 0 F} &
\textbf{16} &
 & \\
\textbf{Average} &
 &
 &
 &
\textbf{1.60} &
 & \\
\midrule

Spreadsheet merge via CLI &
Multi-app &
F & S & 1 &
State Misinterpretation &
Annotation \\

Paper Citation Search &
Multi-app &
F & P & 4 &
Planning Error, Perception Error &
Plan Edit, NL Guidance, Annotation \\

Finding Research Papers &
Multi-app &
F & F & 4 &
Planning Error, Execution Drift &
System-driven IR, Annotation \\

Restaurant Info &
Multi-app &
F & F & 3 &
Planning Error, Execution Drift &
System-driven IR, Annotation \\

PDF Form Filling &
Multi-app &
F & P & 4 &
Execution Drift, State Misinterpretation &
System-driven IR, NL Guidance \\

Revenue + pivot table &
LibreOffice Calc &
F & F & 2 &
Action Grounding Failure, Execution Drift &
Annotation, System-driven IR \\

Branch lookup table &
LibreOffice Calc &
F & P & 2 &
Action Grounding Failure, State Misinterpretation &
Annotation, NL Guidance \\

Calculation + Pivot Table &
LibreOffice Calc &
F & P & 2 &
Missing Context &
NL Guidance \\

Sparkline Charts &
LibreOffice Calc &
F & P & 2 &
Missing Context &
NL Guidance \\

Filling missing data &
LibreOffice Calc &
F & S & 1 &
Planning Error &
Plan Edit \\

Filtering + calculation &
LibreOffice Calc &
F & S & 2 &
Execution Drift &
System-driven IR \\

Text color coding &
LibreOffice Writer &
F & P & 2 &
Planning Error, Execution Drift &
Plan Edit, System-driven IR \\

Font Formatting 1 &
LibreOffice Writer &
F & S & 2 &
Planning Error, Action Grounding &
Plan Edit, Annotation \\

Font Formatting 2 &
LibreOffice Writer &
F & S & 2 &
Execution Drift, State Misinterpretation &
System-driven IR, NL Guidance \\

Locating discussion thread &
Browser &
F & S & 2 &
Execution Drift, Perception Error &
System-driven IR, NL Guidance \\

Hotel booking with filters &
Browser &
F & S & 2 &
Execution Drift, Perception Error &
System-driven IR, Annotation \\

\midrule
\textbf{Total (Auto=F)} &
 &
\textbf{16 F} &
\textbf{7 S / 6 P / 3 F} &
\textbf{37} &
 & \\
\textbf{Average} &
 &
 &
 &
\textbf{2.31} &
 & \\

\midrule
\textbf{Overall Total} &
 &
\textbf{26} &
\textbf{17 S / 6 P / 3 F} &
\textbf{53} &
 & \\
\textbf{Overall Average} &
 &
 &
 &
\textbf{2.04} &
 & \\
\end{tabular}
\end{table*}

%% file: t7_eval_failure_int.tex
\begin{table*}[t]
\centering
\small
\caption{Failure types, recovery mechanisms, and outcomes across 26 non-success trials.}
\label{tab:failure_recovery_summary}

\begin{tabular}{lcccc}
\toprule
\textbf{Category} & \textbf{Count} & \textbf{Success} & \textbf{Partial} & \textbf{Failure} \\
\midrule

\multicolumn{5}{l}{\textit{Failure Types}} \\

Execution Drift & 12 & 8 & 3 & 1 \\
Planning Error & 9 & 6 & 2 & 1 \\
Perception Error & 7 & 5 & 2 & 0 \\
Action Grounding & 5 & 3 & 1 & 1 \\
State Misinterpretation & 5 & 3 & 2 & 0 \\
Missing Context & 2 & 1 & 1 & 0 \\

\midrule

\multicolumn{5}{l}{\textit{Recovery Mechanisms}} \\

System-driven IR & 12 & 7 & 3 & 2 \\
NL Guidance & 12 & 8 & 4 & 0 \\
Annotation & 10 & 7 & 2 & 1 \\
Plan Edit & 6 & 4 & 1 & 1 \\

\bottomrule
\end{tabular}
\end{table*}

%% file: t8_eval_image_comparison.tex
\begin{table*}[t]
\centering
\small
\begin{minipage}{\textwidth} 
    \subsection{Auto-evaluation Details} 
    \vspace{2em} 
\end{minipage}
\caption{Image similarity--based auto-evaluation results for \syss across five scenarios.}
\label{tab:image_similarity_eval}
\begin{tabular}{l p{5.5cm} c c c c c}
\toprule
\textbf{Scenario} & \textbf{Thresholds} & \textbf{Trial} & \textbf{SSIM $\uparrow$} & \textbf{MSE $\downarrow$} & \textbf{dHash $\downarrow$} & \textbf{Match} \\
\midrule
\multirow{6}{*}{Firefox Fillable Form} 
& \multirow{6}{=}{\raggedright
High: $\text{SSIM}\geq0.98$, $\text{MSE}\leq100$, $\text{dHash}\leq2$\\
Partial: $\text{SSIM}\geq0.97$, $\text{MSE}\leq250$, $\text{dHash}\leq5$
}
& 8  & 0.9720 & 205.37 & 5 & Partial \\
& & 10 & 0.9899 & 56.260  & 0 & High \\
& & 11 & 0.9842 & 87.060  & 0 & High \\
& & 41 & 0.9770 & 186.16 & 4 & Partial \\
& & 60 & 0.9856 & 77.320  & 1 & High \\
\cline{3-7}
& & \textbf{Avg} & \textbf{0.9817} & \textbf{122.43} & \textbf{2.0} & -- \\
\midrule

\multirow{6}{*}{Firefox Machine Dashboard} 
& \multirow{6}{=}{\raggedright
High: $\text{SSIM}\geq0.978$, $\text{MSE}\leq140$, $\text{dHash}\leq2$\\
Partial: $\text{SSIM}\geq0.95$, $\text{MSE}\leq250$, $\text{dHash}\leq5$
}
& 5 & 0.9776 & 138.73 & 2 & Partial \\
& & 15 & 0.9825 & 111.55  & 1 & High \\
& & 16 & 0.9796 & 104.82  & 1 & High \\
& & 43 & 0.9763 & 113.06 & 1 & Partial \\
& & 60 & 0.9887 & 59.74  & 1 & High \\
\cline{3-7}
& & \textbf{Avg} & \textbf{0.9809} & \textbf{105.58} & \textbf{1.2} & -- \\
\midrule

\multirow{6}{*}{Firefox Vis Dashboard} 
& \multirow{6}{=}{\raggedright
High: $\text{SSIM}\geq0.97$, $\text{MSE}\leq200$, $\text{dHash}\leq3$\\
Partial: $\text{SSIM}\geq0.90$, $\text{MSE}\leq700$, $\text{dHash}\leq10$
}
& 3 & 0.9168 & 566.26 & 10 & Partial \\
& & 13 & 0.9831 & 106.38  & 1 & High \\
& & 37 & 0.9319 & 582.76  & 8 & Partial \\
& & 64 & 0.9449 & 404.99 & 10 & Partial \\
& & 74 & 0.9164 & 580.11  & 10 & Partial \\
\cline{3-7}
& & \textbf{Avg} & \textbf{0.9386} & \textbf{448.10} & \textbf{7.8} & -- \\
\midrule

\multirow{6}{*}{LibreOffice Incident Sheet} 
& \multirow{6}{=}{\raggedright
High: $\text{SSIM}\geq0.95$, $\text{MSE}\leq2000$, $\text{dHash}\leq5$\\
Partial: $\text{SSIM}\geq0.80$, $\text{MSE}\leq8000$, $\text{dHash}\leq12$
}
& 36 & 0.6448 & 2404.98 & 2 & Low \\
& & 52 & 0.5134 & 2788.43  & 19 & Low \\
& & 66 & 0.7143 & 1461.53  & 18 & Low \\
& & 83 & 0.5720 & 2580.62 & 24 & Low \\
& & 99 & 0.6025 & 1895.99  & 8 & Low \\
\cline{3-7}
& & \textbf{Avg} & \textbf{0.6094} & \textbf{2226.31} & \textbf{14.2} & -- \\
\midrule

\multirow{6}{*}{LibreOffice Sensor Logs Sheet} 
& \multirow{6}{=}{\raggedright
High: $\text{SSIM}\geq0.95$, $\text{MSE}\leq2000$, $\text{dHash}\leq5$\\
Partial: $\text{SSIM}\geq0.80$, $\text{MSE}\leq8000$, $\text{dHash}\leq12$
}
& 6 & 0.6164 & 2347.07 & 14 & Low \\
& & 18 & 0.6383 & 1766.4  & 9 & Low \\
& & 32 & 0.6117 & 2327.08  & 17 & Low \\
& & 57 & 0.9318 & 428.05 & 2 & Partial \\
& & 84 & 0.6292 & 2108.47  & 8 & Low \\
\cline{3-7}
& & \textbf{Avg} & \textbf{0.6855} & \textbf{1795.41} & \textbf{10.0} & -- \\

\midrule
\multicolumn{3}{l}{\textbf{Overall Average}} 
& \textbf{0.8392} 
& \textbf{939.57} 
& \textbf{7.04} 
& -- \\
\bottomrule
\end{tabular}
\end{table*}

%% file: t9_eval_plan_comparison.tex
\begin{table*}[t]
\centering
\small
\caption{Plan comparison metrics across all trials. Coverage measures phase overlap, Order measures phase sequence alignment, Redundancy measures consecutive duplicate phases, and Actionability measures the actionable step ratio. \#Ref refers to the number of reference phases in the random trajectories, while \#Pred refers to the number of reference phases in the predicted trajectories.}
\label{tab:plan_comparison_appendix}

\begin{tabular}{l c c c c c c c}
\toprule
\textbf{Task} & 
\textbf{Trial} & 
\textbf{Coverage} & 
\textbf{Order} & 
\textbf{Redundancy} & 
\textbf{Actionability} & 
\textbf{\#Ref} & 
\textbf{\#Pred} \\
\midrule

\multirow{6}{*}{Firefox Fillable Form}
& 7  & 1.00 & 0.71 & 0.22 & 1.00 & 7  & 7 \\
& 10 & 0.50 & 0.33 & 0.33 & 1.00 & 9  & 6 \\
& 11 & 0.60 & 0.38 & 0.38 & 1.00 & 8  & 5 \\
& 41 & 1.00 & 0.43 & 0.54 & 1.00 & 7  & 6 \\
& 60 & 0.50 & 0.25 & 0.50 & 1.00 & 4  & 3 \\
\cline{2-8}
& \textbf{Avg} & \textbf{0.72} & \textbf{0.42} & \textbf{0.39} & \textbf{1.00} & \textbf{7.0} & \textbf{5.4} \\
\midrule

\multirow{6}{*}{Firefox Machine Dashboard}
& 5  & 0.25 & 0.18 & 0.45 & 1.00 & 11 & 6 \\
& 15 & 0.33 & 0.50 & 0.22 & 1.00 & 4  & 14 \\
& 16 & 0.75 & 0.50 & 0.36 & 1.00 & 10 & 9 \\
& 43 & 0.25 & 0.43 & 0.38 & 0.69 & 7  & 8 \\
& 54 & 0.50 & 0.50 & 0.21 & 1.00 & 8  & 11 \\
\cline{2-8}
& \textbf{Avg} & \textbf{0.42} & \textbf{0.42} & \textbf{0.33} & \textbf{0.94} & \textbf{8.0} & \textbf{9.6} \\
\midrule

\multirow{6}{*}{Firefox Vis Dashboard}
& 3  & 0.75 & 0.38 & 0.25 & 1.00 & 8  & 6 \\
& 13 & 0.67 & 0.38 & 0.13 & 1.00 & 8  & 7 \\
& 37 & 0.50 & 0.42 & 0.25 & 0.83 & 12 & 9 \\
& 64 & 0.67 & 0.64 & 0.23 & 1.00 & 11 & 10 \\
& 74 & 0.67 & 0.40 & 0.19 & 0.88 & 10 & 21 \\
\cline{2-8}
& \textbf{Avg} & \textbf{0.65} & \textbf{0.44} & \textbf{0.21} & \textbf{0.94} & \textbf{9.8} & \textbf{10.6} \\
\midrule

\multirow{6}{*}{LibreOffice Incident Sheet}
& 36 & 0.67 & 0.25 & 0.38 & 1.00 & 8  & 5 \\
& 52 & 0.40 & 0.20 & 0.00 & 1.00 & 10 & 3 \\
& 66 & 0.50 & 0.25 & 0.38 & 1.00 & 4  & 5 \\
& 83 & 1.00 & 0.43 & 0.38 & 1.00 & 7  & 5 \\
& 99 & 0.75 & 0.60 & 0.200 & 1.00 & 5  & 4 \\
\cline{2-8}
& \textbf{Avg} & \textbf{0.66} & \textbf{0.35} & \textbf{0.27} & \textbf{1.00} & \textbf{6.8} & \textbf{4.4} \\
\midrule

\multirow{6}{*}{LibreOffice Sensor Logs}
& 6  & 1.00 & 0.50 & 0.43 & 1.00 & 4 & 4 \\
& 18 & 0.67 & 0.57 & 0.57 & 1.00 & 7 & 6 \\
& 32 & 0.50 & 0.33 & 0.50 & 1.00 & 3 & 3 \\
& 57 & 0.50 & 0.20 & 0.40 & 1.00 & 5 & 3 \\
& 84 & 0.67 & 0.50 & 0.33 & 0.83 & 4 & 8 \\
\cline{2-8}
& \textbf{Avg} & \textbf{0.67} & \textbf{0.42} & \textbf{0.44} & \textbf{0.97} & \textbf{4.6} & \textbf{4.8} \\
\midrule

\textbf{Overall Avg} & -- & \textbf{0.62} & \textbf{0.41} & \textbf{0.33} & \textbf{0.97} & \textbf{7.24} & \textbf{6.96} \\
\bottomrule
\end{tabular}
\end{table*}

%% file: main.bib
@inproceedings{
wang2026computer,
title={Computer Agent Arena: Toward Human-Centric Evaluation and Analysis of Computer-Use Agents},
author={Bowen Wang and Xinyuan Wang and Jiaqi Deng and Tianbao Xie and Ryan Li and Yanzhe Zhang and Junli Wang and Dunjie Lu and Zicheng Gong and Gavin Li and Toh Jing Hua and Wei-Lin Chiang and Ion Stoica and Diyi Yang and Yu Su and Yi Zhang and Zhiguo Wang and Victor Zhong and Tao Yu},
booktitle={The Fourteenth International Conference on Learning Representations},
year={2026},
url={https://openreview.net/forum?id=3x4SDbXbgl}
}

@misc{
lu2024omniparserpurevisionbased,
title={OmniParser for Pure Vision Based {GUI} Agent},
author={Yadong Lu and Jianwei Yang and Yelong Shen and Ahmed Hassan Awadallah},
year={2025},
url={https://openreview.net/forum?id=C6hUK6Q1Pi}
}

@inproceedings{chen2025interchat,
  title={InterChat: Enhancing generative visual analytics using multimodal interactions},
  author={Chen, Juntong and Wu, Jiang and Guo, Jiajing and Mohanty, Vikram and Li, Xueming and Ono, Jorge Piazentin and He, Wenbin and Ren, Liu and Liu, Dongyu},
  booktitle={Computer Graphics Forum},
  volume={44},
  number={3},
  pages={e70112},
  year={2025},
  organization={Wiley Online Library}
}

@inproceedings{hellmann2011rule,
  title={Rule-based exploratory testing of graphical user interfaces},
  author={Hellmann, Theodore D and Maurer, Frank},
  booktitle={2011 Agile Conference},
  pages={107--116},
  year={2011},
  organization={IEEE}
}

@inproceedings{qian2020roscript,
  title={Roscript: a visual script driven truly non-intrusive robotic testing system for touch screen applications},
  author={Qian, Ju and Shang, Zhengyu and Yan, Shuoyan and Wang, Yan and Chen, Lin},
  booktitle={Proceedings of the ACM/IEEE 42nd International Conference on Software Engineering},
  pages={297--308},
  year={2020}
}

@inproceedings{granda2021towards,
  title={Towards a Model-Driven Testing Framework for GUI Test Cases Generation from User Stories.},
  author={Granda, Maria Fernanda and Parra, Otto and Alba-Sarango, Bryan},
  booktitle={ENASE},
  pages={453--460},
  year={2021}
}

@article{hofmann2020robotic,
  title={Robotic process automation},
  author={Hofmann, Peter and Samp, Caroline and Urbach, Nils},
  journal={Electronic markets},
  volume={30},
  number={1},
  pages={99--106},
  year={2020},
  publisher={Springer}
}

@article{pruucha2025llm,
  title={Are LLM Agents the New RPA? A Comparative Study with RPA Across Enterprise Workflows},
  author = {Petr Pr{\r{u}}cha and Michaela Matou{\v{s}}kov{\'a} and Jan Strnad},
  journal={arXiv preprint arXiv:2509.04198},
  year={2025}
}

@article{wewerka2020robotic,
  author       = {Judith Wewerka and
                  Manfred Reichert},
  title        = {Robotic Process Automation - {A} Systematic Literature Review and
                  Assessment Framework},
  journal      = {CoRR},
  volume       = {abs/2012.11951},
  year         = {2020},
  url          = {https://arxiv.org/abs/2012.11951},
  eprinttype    = {arXiv},
  eprint       = {2012.11951}
}

@article{dwarakanath2018machines,
  author       = {Anurag Dwarakanath and
                  Neville Dubash and
                  Sanjay Podder},
  title        = {Machines that test Software like Humans},
  journal      = {CoRR},
  volume       = {abs/1809.09455},
  year         = {2018},
  url          = {http://arxiv.org/abs/1809.09455},
  eprinttype    = {arXiv},
  eprint       = {1809.09455}
}

@misc{jain2024smartflow,
      title={SmartFlow: Robotic Process Automation using LLMs}, 
      author={Arushi Jain and Shubham Paliwal and Monika Sharma and Lovekesh Vig and Gautam Shroff},
      year={2024},
      eprint={2405.12842},
      archivePrefix={arXiv},
      primaryClass={cs.RO},
      url={https://arxiv.org/abs/2405.12842}, 
}

@inproceedings{coppola2019fragility, 
   title={Fragility of layout-based and visual GUI test scripts: an assessment study on a hybrid mobile application},
   url={http://dx.doi.org/10.1145/3340433.3342824},
   DOI={10.1145/3340433.3342824},
   booktitle={Proceedings of the 10th ACM SIGSOFT International Workshop on Automating TEST Case Design, Selection, and Evaluation},
   publisher={ACM},
   author={Coppola, Riccardo and Ardito, Luca and Torchiano, Marco},
   year={2019},
   month=Aug, pages={28–34},
   collection={ESEC/FSE ’19} }

@article{brisset2022erratum,
  author       = {Sacha Brisset and
                  Romain Rouvoy and
                  Lionel Seinturier and
                  Renaud Pawlak},
  title        = {Erratum: Leveraging Flexible Tree Matching to repair broken locators
                  in web automation scripts},
  journal      = {Inf. Softw. Technol.},
  volume       = {144},
  pages        = {106754},
  year         = {2022},
  url          = {https://doi.org/10.1016/j.infsof.2021.106754},
  doi          = {10.1016/J.INFSOF.2021.106754}
}

@inproceedings{song2025can,
  author       = {Zihe Song and
                  S. M. Hasan Mansur and
                  Ravishka Rathnasuriya and
                  Yumna Fatima and
                  Wei Yang and
                  Kevin Moran and
                  Wing Lam},
  title        = {Can You Mimic Me? Exploring the Use of Android Record {\&} Replay
                  Tools in Debugging},
  booktitle    = {12th {IEEE/ACM} International Conference on Mobile Software Engineering
                  and Systems, MOBILESoft@ICSE 2025, Ottawa, ON, Canada, April 27-28,
                  2025},
  pages        = {32--43},
  publisher    = {{IEEE}},
  year         = {2025},
  url          = {https://doi.org/10.1109/MOBILESoft66462.2025.00011},
  doi          = {10.1109/MOBILESOFT66462.2025.00011}
}

@inproceedings{yu2024practical,
  author       = {Shengcheng Yu and
                  Chunrong Fang and
                  Mingzhe Du and
                  Yuchen Ling and
                  Zhenyu Chen and
                  Zhendong Su},
  title        = {Practical Non-Intrusive {GUI} Exploration Testing with Visual-based
                  Robotic Arms},
  booktitle    = {Proceedings of the 46th {IEEE/ACM} International Conference on Software
                  Engineering, {ICSE} 2024, Lisbon, Portugal, April 14-20, 2024},
  pages        = {130:1--130:13},
  publisher    = {{ACM}},
  year         = {2024},
  url          = {https://doi.org/10.1145/3597503.3639161},
  doi          = {10.1145/3597503.3639161}
}

@inproceedings{yeh2009sikuli,
  author       = {Tom Yeh and
                  Tsung{-}Hsiang Chang and
                  Robert C. Miller},
  editor       = {Andrew D. Wilson and
                  Fran{\c{c}}ois Guimbreti{\`{e}}re},
  title        = {Sikuli: using {GUI} screenshots for search and automation},
  booktitle    = {Proceedings of the 22nd Annual {ACM} Symposium on User Interface Software
                  and Technology, Victoria, BC, Canada, October 4-7, 2009},
  pages        = {183--192},
  publisher    = {{ACM}},
  year         = {2009},
  url          = {https://doi.org/10.1145/1622176.1622213},
  doi          = {10.1145/1622176.1622213}
}

@inproceedings{nguyen2025gui,
  author       = {Dang Nguyen and
                  Jian Chen and
                  Yu Wang and
                  Gang Wu and
                  Namyong Park and
                  Zhengmian Hu and
                  Hanjia Lyu and
                  Junda Wu and
                  Ryan Aponte and
                  Yu Xia and
                  Xintong Li and
                  Jing Shi and
                  Hongjie Chen and
                  Viet Dac Lai and
                  Zhouhang Xie and
                  Sungchul Kim and
                  Ruiyi Zhang and
                  Tong Yu and
                  Md. Mehrab Tanjim and
                  Nesreen K. Ahmed and
                  Puneet Mathur and
                  Seunghyun Yoon and
                  Lina Yao and
                  Branislav Kveton and
                  Jihyung Kil and
                  Thien Huu Nguyen and
                  Trung Bui and
                  Tianyi Zhou and
                  Ryan A. Rossi and
                  Franck Dernoncourt},
  editor       = {Wanxiang Che and
                  Joyce Nabende and
                  Ekaterina Shutova and
                  Mohammad Taher Pilehvar},
  title        = {{GUI} Agents: {A} Survey},
  booktitle    = {Findings of the Association for Computational Linguistics, {ACL} 2025,
                  Vienna, Austria, July 27 - August 1, 2025},
  series       = {Findings of {ACL}},
  volume       = {{ACL} 2025},
  pages        = {22522--22538},
  publisher    = {Association for Computational Linguistics},
  year         = {2025},
  url          = {https://aclanthology.org/2025.findings-acl.1158/}
}

@misc{zhao2025osagentssurvey,
      title={OS Agents: A Survey on MLLM-based Agents for General Computing Devices Use},
      author={Zhou Zhao and Shengyu Zhang and Liang Wang and Xiangxin Zhou and Zhaokai Wang and Kun Kuang and Fei Wu and Wangchunshu Zhou and Shuofei Qiao and Jiwei Li and Guoyin Wang and Ziyu Zhao and Hongxia Yang and Fan Wu and Jiasheng Ye and Shenzhi Wang and Ruixuan Xiao and Tieyong Zeng and Yuhuai Li and Yuchen Eleanor Jiang and Meiling Tao and Xueyu Hu and Tao Xiong and Biao Yi and Keting Yin and Yurun Chen and Zishu Wei and Xinchen Xu and Shengze Xu},
      year={2025},
      eprint={2508.04482},
      archivePrefix={arXiv},
      primaryClass={cs.AI},
      url={https://arxiv.org/abs/2508.04482},
}

@article{tang2025survey,
  title={A survey on (m) llm-based gui agents},
  author={Tang, Fei and Xu, Haolei and Zhang, Hang and Chen, Siqi and Wu, Xingyu and Shen, Yongliang and Zhang, Wenqi and Hou, Guiyang and Tan, Zeqi and Yan, Yuchen and others},
  journal={arXiv preprint arXiv:2504.13865},
  year={2025}
}

@article{wu2024visual,
  author       = {Junda Wu and
                  Zhehao Zhang and
                  Yu Xia and
                  Xintong Li and
                  Zhaoyang Xia and
                  Aaron Chang and
                  Tong Yu and
                  Sungchul Kim and
                  Ryan A. Rossi and
                  Ruiyi Zhang and
                  Subrata Mitra and
                  Dimitris N. Metaxas and
                  Lina Yao and
                  Jingbo Shang and
                  Julian J. McAuley},
  title        = {Visual Prompting in Multimodal Large Language Models: {A} Survey},
  journal      = {CoRR},
  volume       = {abs/2409.15310},
  year         = {2024},
  url          = {https://doi.org/10.48550/arXiv.2409.15310},
  doi          = {10.48550/ARXIV.2409.15310},
  eprinttype    = {arXiv},
  eprint       = {2409.15310}
}

@inproceedings{chen2023gap,
  title={From gap to synergy: Enhancing contextual understanding through human-machine collaboration in personalized systems},
  author={Chen, Weihao and Yu, Chun and Wang, Huadong and Wang, Zheng and Yang, Lichen and Wang, Yukun and Shi, Weinan and Shi, Yuanchun},
  booktitle={Proceedings of the 36th Annual ACM Symposium on User Interface Software and Technology},
  pages={1--15},
  year={2023}
}

@article{xu2025duetui,
  author       = {Yuan Xu and
                  Shaowen Xiang and
                  Yizhi Song and
                  Ruoting Sun and
                  Xin Tong},
  editor       = {Nuria Oliver and
                  David A. Shamma and
                  Heloisa Candello and
                  Pablo C{\'{e}}sar and
                  Pedro Lopes and
                  Alessandro Bozzon and
                  Thomas Kosch and
                  Vera Q. Liao and
                  Xiaojuan Ma and
                  Valentino Artizzu and
                  Fiona Draxler and
                  Gustavo L{\'{o}}pez and
                  Anke V. Reinschluessel and
                  Xin Tong and
                  Phoebe O. Toups Dugas},
  title        = {DuetUI: {A} Bidirectional Context Loop for Human-Agent Co-Generation
                  of Task-Oriented Interfaces},
  journal    = {Proceedings of the 2026 {CHI} Conference on Human Factors in Computing
                  Systems, {CHI} 2026, Barcelona, Spain, April 13-17, 2026},
  pages        = {305:1--305:22},
  publisher    = {ACM},
  year         = {2026}
}

@inproceedings{goyal2024designing,
  title={Designing for Human-Agent Alignment: Understanding what humans want from their agents},
  author={Goyal, Nitesh and Chang, Minsuk and Terry, Michael},
  booktitle={Extended Abstracts of the CHI Conference on Human Factors in Computing Systems},
  pages={1--6},
  year={2024}
}

@misc{issak2025mosaaic,
      title={MOSAAIC: Managing Optimization towards Shared Autonomy, Authority, and Initiative in Co-creation}, 
      author={Alayt Issak and Jeba Rezwana and Casper Harteveld},
      year={2025},
      eprint={2505.11481},
      archivePrefix={arXiv},
      primaryClass={cs.AI},
      url={https://arxiv.org/abs/2505.11481}, 
}

@article{tang2025dark,
title={Dark patterns meet gui agents: Llm agent susceptibility to manipulative interfaces and the role of human oversight},
  author={Tang, Jingyu and Chen, Chaoran and Li, Jiawen and Zhang, Zhiping and Guo, Bingcan and Khalilov, Ibrahim and Gebreegziabher, Simret Araya and Yao, Bingsheng and Wang, Dakuo and Ye, Yanfang and others},
  journal={Proceedings of the 2026 CHI Conference on Human Factors in Computing Systems},
  pages={1--26},
  year={2026}
}

@article{rabanser2026towards,
  title={Towards a Science of AI Agent Reliability},
  author={Rabanser, Stephan and Kapoor, Sayash and Kirgis, Peter and Liu, Kangheng and Utpala, Saiteja and Narayanan, Arvind},
  journal={arXiv preprint arXiv:2602.16666},
  year={2026}
}

@inproceedings{chakraborti2019plan,
  title={Plan explanations as model reconciliation--an empirical study},
  author={Chakraborti, Tathagata and Sreedharan, Sarath and Grover, Sachin and Kambhampati, Subbarao},
  booktitle={2019 14th ACM/IEEE International Conference on Human-Robot Interaction (HRI)},
  pages={258--266},
  year={2019},
  organization={Ieee}
}

@article{vats2024survey,
  title={A Survey on Human-AI Collaboration with Large Foundation Models},
  author={Vats, Vanshika and Nizam, Marzia Binta and Liu, Minghao and Wang, Ziyuan and Ho, Richard and Prasad, Mohnish Sai and Titterton, Vincent and Malreddy, Sai Venkat and Aggarwal, Riya and Xu, Yanwen and others},
  journal={arXiv preprint arXiv:2403.04931},
  year={2024}
}

@article{zou2025llm,
  title={Llm-based human-agent collaboration and interaction systems: A survey},
  author={Zou, Henry Peng and Huang, Wei-Chieh and Wu, Yaozu and Guo, Jizhou and Chen, Yankai and Miao, Chunyu and Nguyen, Hoang H and Zhou, Yue and Zhang, Weizhi and Fang, Liancheng and others},
  journal={Findings of the Association for Computational Linguistics: ACL 2026},
  pages={36335--36364},
  year={2026}
}

@article{potts2026invisible,
  title={Invisible failures in human-AI interactions},
  author={Potts, Christopher and Sudhof, Moritz},
  journal={arXiv preprint arXiv:2603.15423},
  year={2026}
}

@inproceedings{horvitz1999principles,
  author       = {Eric Horvitz},
  editor       = {Marian G. Williams and
                  Mark W. Altom},
  title        = {Principles of Mixed-Initiative User Interfaces},
  booktitle    = {Proceeding of the {CHI} '99 Conference on Human Factors in Computing
                  Systems: The {CHI} is the Limit, Pittsburgh, PA, USA, May 15-20, 1999},
  pages        = {159--166},
  publisher    = {{ACM}},
  year         = {1999},
  url          = {https://doi.org/10.1145/302979.303030},
  doi          = {10.1145/302979.303030}
}

@article{son2026hand,
  title={" When to Hand Off, When to Work Together": Expanding Human-Agent Co-Creative Collaboration through Concurrent Interaction},
  author={Son, Kihoon and Lee, Hyewon and Choi, DaEun and Kim, Yoonsu and Kim, Tae Soo and Lee, Yoonjoo and Chung, John Joon Young and Jung, HyunJoon and Kim, Juho},
  journal={arXiv preprint arXiv:2603.02050},
  year={2026}
}

@inproceedings{chakraborti2017visualizations,
  author       = {Tathagata Chakraborti and
                  Kshitij P. Fadnis and
                  Kartik Talamadupula and
                  Mishal Dholakia and
                  Biplav Srivastava and
                  Jeffrey O. Kephart and
                  Rachel K. E. Bellamy},
  editor       = {J{\'{e}}r{\^{o}}me Lang},
  title        = {Visualizations for an Explainable Planning Agent},
  booktitle    = {Proceedings of the Twenty-Seventh International Joint Conference on
                  Artificial Intelligence, {IJCAI} 2018, July 13-19, 2018, Stockholm,
                  Sweden},
  pages        = {5820--5822},
  publisher    = {ijcai.org},
  year         = {2018}
}

@article{narechania2025utilizing,
  author       = {Arpit Narechania and
                  Shunan Guo and
                  Eunyee Koh and
                  Alex Endert and
                  Jane Hoffswell},
  title        = {Utilizing Provenance as an Attribute for Visual Data Analysis: {A}
                  Design Probe With ProvenanceLens},
  journal      = {{IEEE} Trans. Vis. Comput. Graph.},
  volume       = {31},
  number       = {10},
  pages        = {8452--8465},
  year         = {2025}
}

@inproceedings{nayak2025ui,
  title = 	 {{UI}-Vision: A Desktop-centric {GUI} Benchmark for Visual Perception and Interaction},
  author =       {Nayak, Shravan and Jian, Xiangru and Lin, Kevin Qinghong and Rodriguez, Juan A. and Kalsi, Montek and Chapados, Nicolas and \"{O}zsu, M. Tamer and Agrawal, Aishwarya and Vazquez, David and Pal, Christopher and Taslakian, Perouz and Gella, Spandana and Rajeswar, Sai},
  booktitle = 	 {Proceedings of the 42nd International Conference on Machine Learning},
  pages = 	 {45817--45851},
  year = 	 {2025},
  editor = 	 {Singh, Aarti and Fazel, Maryam and Hsu, Daniel and Lacoste-Julien, Simon and Berkenkamp, Felix and Maharaj, Tegan and Wagstaff, Kiri and Zhu, Jerry},
  volume = 	 {267},
  series = 	 {Proceedings of Machine Learning Research},
  month = 	 {13--19 Jul},
  publisher =    {PMLR},
}

@inproceedings{
shayegani2025just,
title={Just Do It!? Computer-Use Agents Exhibit Blind Goal-Directedness},
author={Erfan Shayegani and Keegan Hines and Yue Dong and Nael Abu-Ghazaleh and Roman Lutz and Spencer Whitehead and Vidhisha Balachandran and Besmira Nushi and Vibhav Vineet},
booktitle={The Fourteenth International Conference on Learning Representations},
year={2026},
url={https://openreview.net/forum?id=9W4bPRsEIT}
}

@article{shaw2023pixels,
  title={From pixels to ui actions: Learning to follow instructions via graphical user interfaces},
  author={Shaw, Peter and Joshi, Mandar and Cohan, James and Berant, Jonathan and Pasupat, Panupong and Hu, Hexiang and Khandelwal, Urvashi and Lee, Kenton and Toutanova, Kristina N},
  journal={Advances in Neural Information Processing Systems},
  volume={36},
  pages={34354--34370},
  year={2023}
}

@inproceedings{gou2024navigating,
    title={Navigating the digital world as humans do: Universal visual grounding for gui agents},
  author={Gou, Boyu and Wang, Demi Ruohan and Zheng, Boyuan and Xie, Yanan and Chang, Cheng and Shu, Yiheng and Sun, Huan and Su, Yu},
  booktitle={International Conference on Learning Representations},
  volume={2025},
  pages={30851--30883},
  year={2025}
}

@inproceedings{lin2025showui,
  title={Showui: One vision-language-action model for gui visual agent},
  author={Lin, Kevin Qinghong and Li, Linjie and Gao, Difei and Yang, Zhengyuan and Wu, Shiwei and Bai, Zechen and Lei, Stan Weixian and Wang, Lijuan and Shou, Mike Zheng},
  booktitle={Proceedings of the Computer Vision and Pattern Recognition Conference},
  pages={19498--19508},
  year={2025}
}

@article{bhathal2025websight,
  title={Websight: A vision-first architecture for robust web agents},
  author={Bhathal, Tanvir and Gupta, Asanshay},
  journal={arXiv preprint arXiv:2508.16987},
  year={2025}
}

@inproceedings{huang2025spiritsight,
  title={Spiritsight agent: Advanced gui agent with one look},
  author={Huang, Zhiyuan and Cheng, Ziming and Pan, Junting and Hou, Zhaohui and Zhan, Mingjie},
  booktitle={Proceedings of the Computer Vision and Pattern Recognition Conference},
  pages={29490--29500},
  year={2025}
}

@article{yang2025ultracua,
  title={Ultracua: A foundation model for computer use agents with hybrid action},
  author={Yang, Yuhao and Yang, Zhen and Dou, Zi-Yi and Nguyen, Anh and You, Keen and Attia, Omar and Szot, Andrew and Feng, Michael and Ramrakhya, Ram and Toshev, Alexander and others},
  journal={arXiv preprint arXiv:2510.17790},
  year={2025}
}

@article{zhang2025btl,
  title={Btl-ui: Blink-think-link reasoning model for gui agent},
  author={Zhang, Shaojie and Zhang, Ruoceng and Fu, Pei and Wang, Shaokang and Yang, Jiahui and Du, Xin and Qin, Bin and Huang, Ying and Luo, Zhenbo and Luan, Jian},
  journal={Advances in Neural Information Processing Systems},
  volume={38},
  pages={56035--56056},
  year={2026}
}

@inproceedings{he2024webvoyager,
  title={Webvoyager: Building an end-to-end web agent with large multimodal models},
  author={He, Hongliang and Yao, Wenlin and Ma, Kaixin and Yu, Wenhao and Dai, Yong and Zhang, Hongming and Lan, Zhenzhong and Yu, Dong},
  booktitle={Proceedings of the 62nd Annual Meeting of the Association for Computational Linguistics (Volume 1: Long Papers)},
  pages={6864--6890},
  year={2024}
}

@inproceedings{song2024visiontasker,
  title={Visiontasker: Mobile task automation using vision based ui understanding and llm task planning},
  author={Song, Yunpeng and Bian, Yiheng and Tang, Yongtao and Ma, Guiyu and Cai, Zhongmin},
  booktitle={Proceedings of the 37th Annual ACM Symposium on User Interface Software and Technology},
  pages={1--17},
  year={2024}
}

@inproceedings{wu2024atlas,
  author       = {Zhiyong Wu and
                  Zhenyu Wu and
                  Fangzhi Xu and
                  Yian Wang and
                  Qiushi Sun and
                  Chengyou Jia and
                  Kanzhi Cheng and
                  Zichen Ding and
                  Liheng Chen and
                  Paul Pu Liang and
                  Yu Qiao},
  title        = {{OS-ATLAS:} Foundation Action Model for Generalist {GUI} Agents},
  booktitle    = {The Thirteenth International Conference on Learning Representations,
                  {ICLR} 2025, Singapore, April 24-28, 2025},
  publisher    = {OpenReview.net},
  year         = {2025},
  url          = {https://openreview.net/forum?id=n9PDaFNi8t}
}

@article{qin2025ui,
  author       = {Yujia Qin and
                  Yining Ye and
                  Junjie Fang and
                  Haoming Wang and
                  Shihao Liang and
                  Shizuo Tian and
                  Junda Zhang and
                  Jiahao Li and
                  Yunxin Li and
                  Shijue Huang and
                  Wanjun Zhong and
                  Kuanye Li and
                  Jiale Yang and
                  Yu Miao and
                  Woyu Lin and
                  Longxiang Liu and
                  Xu Jiang and
                  Qianli Ma and
                  Jingyu Li and
                  Xiaojun Xiao and
                  Kai Cai and
                  Chuang Li and
                  Yaowei Zheng and
                  Chaolin Jin and
                  Chen Li and
                  Xiao Zhou and
                  Minchao Wang and
                  Haoli Chen and
                  Zhaojian Li and
                  Haihua Yang and
                  Haifeng Liu and
                  Feng Lin and
                  Tao Peng and
                  Xin Liu and
                  Guang Shi},
  title        = {{UI-TARS:} Pioneering Automated {GUI} Interaction with Native Agents},
  journal      = {CoRR},
  volume       = {abs/2501.12326},
  year         = {2025},
  url          = {https://doi.org/10.48550/arXiv.2501.12326},
  doi          = {10.48550/ARXIV.2501.12326},
  eprinttype    = {arXiv},
  eprint       = {2501.12326}
}

@article{drouin2024workarena,
  author       = {Alexandre Drouin and
                  Maxime Gasse and
                  Massimo Caccia and
                  Issam H. Laradji and
                  Manuel Del Verme and
                  Tom Marty and
                  David V{\'{a}}zquez and
                  Nicolas Chapados and
                  Alexandre Lacoste},
  editor       = {Ruslan Salakhutdinov and
                  Zico Kolter and
                  Katherine A. Heller and
                  Adrian Weller and
                  Nuria Oliver and
                  Jonathan Scarlett and
                  Felix Berkenkamp},
  title        = {WorkArena: How Capable are Web Agents at Solving Common Knowledge
                  Work Tasks?},
  journal    = {Forty-first International Conference on Machine Learning, {ICML} 2024,
                  Vienna, Austria, July 21-27, 2024},
  series       = {Proceedings of Machine Learning Research},
  volume       = {235},
  pages        = {11642--11662},
  publisher    = {{PMLR} / OpenReview.net},
  year         = {2024}
}

@article{boisvert2024workarena++,
  title={Workarena++: Towards compositional planning and reasoning-based common knowledge work tasks},
  author={Boisvert, L{\'e}o and Thakkar, Megh and Gasse, Maxime and Caccia, Massimo and De Chezelles, Thibault L and Cappart, Quentin and Chapados, Nicolas and Lacoste, Alexandre and Drouin, Alexandre},
  journal={Advances in Neural Information Processing Systems},
  volume={37},
  pages={5996--6051},
  year={2024}
}

@inproceedings{koh2024visualwebarena,
  title={Visualwebarena: Evaluating multimodal agents on realistic visual web tasks},
  author={Koh, Jing Yu and Lo, Robert and Jang, Lawrence and Duvvur, Vikram and Lim, Ming and Huang, Po-Yu and Neubig, Graham and Zhou, Shuyan and Salakhutdinov, Russ and Fried, Daniel},
  booktitle={Proceedings of the 62nd Annual Meeting of the Association for Computational Linguistics (Volume 1: Long Papers)},
  pages={881--905},
  year={2024}
}

@inproceedings{zhou2023webarena,
  title={Webarena: A realistic web environment for building autonomous agents},
  author={Zhou, Shuyan and Xu, Frank F and Zhu, Hao and Zhou, Xuhui and Lo, Robert and Sridhar, Abishek and Cheng, Xianyi and Ou, Tianyue and Bisk, Yonatan and Fried, Daniel and others},
  booktitle={International Conference on Learning Representations},
  volume={2024},
  pages={15585--15606},
  year={2024}
}

@article{yao2022webshop,
  title={Webshop: Towards scalable real-world web interaction with grounded language agents},
  author={Yao, Shunyu and Chen, Howard and Yang, John and Narasimhan, Karthik},
  journal={Advances in Neural Information Processing Systems},
  volume={35},
  pages={20744--20757},
  year={2022}
}

@article{yang2024swe,
  title={Swe-agent: Agent-computer interfaces enable automated software engineering},
  author={Yang, John and Jimenez, Carlos E and Wettig, Alexander and Lieret, Kilian and Yao, Shunyu and Narasimhan, Karthik and Press, Ofir},
  journal={Advances in Neural Information Processing Systems},
  volume={37},
  pages={50528--50652},
  year={2024}
}

@article{hu2024dawn,
  title={The dawn of gui agent: A preliminary case study with claude 3.5 computer use},
  author={Hu, Siyuan and Ouyang, Mingyu and Gao, Difei and Shou, Mike Zheng},
  journal={arXiv preprint arXiv:2411.10323},
  year={2024}
}

@inproceedings{xie2024osworld,
  author       = {Tianbao Xie and
                  Danyang Zhang and
                  Jixuan Chen and
                  Xiaochuan Li and
                  Siheng Zhao and
                  Ruisheng Cao and
                  Toh Jing Hua and
                  Zhoujun Cheng and
                  Dongchan Shin and
                  Fangyu Lei and
                  Yitao Liu and
                  Yiheng Xu and
                  Shuyan Zhou and
                  Silvio Savarese and
                  Caiming Xiong and
                  Victor Zhong and
                  Tao Yu},
  editor       = {Amir Globersons and
                  Lester Mackey and
                  Danielle Belgrave and
                  Angela Fan and
                  Ulrich Paquet and
                  Jakub M. Tomczak and
                  Cheng Zhang},
  title        = {OSWorld: Benchmarking Multimodal Agents for Open-Ended Tasks in Real
                  Computer Environments},
  booktitle    = {Advances in Neural Information Processing Systems 38: Annual Conference
                  on Neural Information Processing Systems 2024, NeurIPS 2024, Vancouver,
                  BC, Canada, December 10 - 15, 2024},
  year         = {2024}
}

@inproceedings{yao2022react,
  title={React: Synergizing reasoning and acting in language models},
  author={Yao, Shunyu and Zhao, Jeffrey and Yu, Dian and Du, Nan and Shafran, Izhak and Narasimhan, Karthik R and Cao, Yuan},
  booktitle={The eleventh international conference on learning representations},
  year={2022}
}

@article{team2026kimi,
  title={Kimi K2. 5: Visual Agentic Intelligence},
  author={Team, Kimi and Bai, Tongtong and Bai, Yifan and Bao, Yiping and Cai, SH and Cao, Yuan and Charles, Y and Che, HS and Chen, Cheng and Chen, Guanduo and others},
  journal={arXiv preprint arXiv:2602.02276},
  year={2026}
}

@article{shen2025mind,
  title={From mind to machine: The rise of manus ai as a fully autonomous digital agent},
  author={Shen, Minjie and Li, Yanshu and Chen, Lulu and Yang, Qikai},
  journal={arXiv preprint arXiv:2505.02024},
  year={2025}
}

@inproceedings{wen2024autodroid,
  title={Autodroid: Llm-powered task automation in android},
  author={Wen, Hao and Li, Yuanchun and Liu, Guohong and Zhao, Shanhui and Yu, Tao and Li, Toby Jia-Jun and Jiang, Shiqi and Liu, Yunhao and Zhang, Yaqin and Liu, Yunxin},
  booktitle={Proceedings of the 30th Annual International Conference on Mobile Computing and Networking},
  pages={543--557},
  year={2024}
}

@article{ma2025agent+,
  title={Agent+ P: Guiding UI Agents via Symbolic Planning},
  author={Ma, Shang and Xiao, Xusheng and Ye, Yanfang},
  journal={arXiv preprint arXiv:2510.06042},
  year={2025}
}

@inproceedings{erdogan2025plan,
  title = 	 {Plan-and-Act: Improving Planning of Agents for Long-Horizon Tasks},
  author =       {Erdogan, Lutfi Eren and Lee, Nicholas and Kim, Sehoon and Moon, Suhong and Furuta, Hiroki and Anumanchipalli, Gopala and Keutzer, Kurt and Gholami, Amir},
  booktitle = 	 {Proceedings of the 42nd International Conference on Machine Learning},
  pages = 	 {15419--15462},
  year = 	 {2025},
  editor = 	 {Singh, Aarti and Fazel, Maryam and Hsu, Daniel and Lacoste-Julien, Simon and Berkenkamp, Felix and Maharaj, Tegan and Wagstaff, Kiri and Zhu, Jerry},
  volume = 	 {267},
  series = 	 {Proceedings of Machine Learning Research},
  month = 	 {13--19 Jul},
  publisher =    {PMLR}
}

@article{zhang2026showui,
  author       = {Yichun Zhang and
                  Xiangwu Guo and
                  Yauhong Goh and
                  Jessica Hu and
                  Zhiheng Chen and
                  Xin Wang and
                  Difei Gao and
                  Mike Zheng Shou},
  title        = {ShowUI-Aloha: Human-Taught {GUI} Agent},
  journal      = {CoRR},
  volume       = {abs/2601.07181},
  year         = {2026}
}

@inproceedings{aghzal2026llm,
   title={Why do LLM-based web agents fail? A hierarchical planning perspective},
  author={Aghzal, Mohamed and Stein, Gregory J and Yao, Ziyu},
  booktitle={Proceedings of the 64th Annual Meeting of the Association for Computational Linguistics (Volume 1: Long Papers)},
  pages={32157--32180},
  year={2026}
}

@inproceedings{hua2025interactive,
   title={Interactive speculative planning: Enhance agent efficiency through co-design of system and user interface},
  author={Hua, Wenyue and Wan, Mengting and Vadrevu, Jagannath and Nadel, Ryan and Zhang, Yongfeng and Wang, Chi},
  booktitle={International Conference on Learning Representations},
  volume={2025},
  pages={14256--14283},
  year={2025}
}

@inproceedings{li2017sugilite,
  title={SUGILITE: creating multimodal smartphone automation by demonstration},
  author={Li, Toby Jia-Jun and Azaria, Amos and Myers, Brad A},
  booktitle={Proceedings of the 2017 CHI conference on human factors in computing systems},
  pages={6038--6049},
  year={2017}
}

@inproceedings{khurana2025me,
  author       = {Anjali Khurana and
                  Xiaotian Su and
                  April Yi Wang and
                  Parmit K. Chilana},
  editor       = {Naomi Yamashita and
                  Vanessa Evers and
                  Koji Yatani and
                  Sharon Xianghua Ding and
                  Bongshin Lee and
                  Marshini Chetty and
                  Phoebe O. Toups Dugas},
  title        = {Do It For Me vs. Do It With Me: Investigating User Perceptions of
                  Different Paradigms of Automation in Copilots for Feature-Rich Software},
  booktitle    = {Proceedings of the 2025 {CHI} Conference on Human Factors in Computing
                  Systems, {CHI} 2025, YokohamaJapan, 26 April 2025- 1 May 2025},
  pages        = {880:1--880:18},
  publisher    = {{ACM}},
  year         = {2025},
  url          = {https://doi.org/10.1145/3706598.3713431},
  doi          = {10.1145/3706598.3713431}
}

@inproceedings{amershi2019guidelines,
  title={Guidelines for human-AI interaction},
  author={Amershi, Saleema and Weld, Dan and Vorvoreanu, Mihaela and Fourney, Adam and Nushi, Besmira and Collisson, Penny and Suh, Jina and Iqbal, Shamsi and Bennett, Paul N and Inkpen, Kori and others},
  booktitle={Proceedings of the 2019 CHI Conference on Human Factors in Computing Systems},
  pages={1--13},
  year={2019}
}

@inproceedings{feng2024cocoa,
  title={Cocoa: Co-planning and co-execution with ai agents},
  author={Feng, KJ Kevin and Pu, Kevin and Latzke, Matt and August, Tal and Siangliulue, Pao and Bragg, Jonathan and Weld, Daniel S and Zhang, Amy X and Chang, Joseph Chee},
  booktitle={Proceedings of the 2026 CHI Conference on Human Factors in Computing Systems},
  pages={1--23},
  year={2026},
  publisher    = {{ACM}}
}

@inproceedings{huq2025cowpilot,
   title={Cowpilot: a framework for autonomous and human-agent collaborative web navigation},
  author={Huq, Faria and Wang, Zora Zhiruo and Xu, Frank F and Ou, Tianyue and Zhou, Shuyan and Bigham, Jeffrey P and Neubig, Graham},
  booktitle={Proceedings of the 2025 Conference of the Nations of the Americas Chapter of the Association for Computational Linguistics: Human Language Technologies (System Demonstrations)},
  pages={163--172},
  year={2025}
}

@article{mozannar2025magentic,
  title={Magentic-ui: Towards human-in-the-loop agentic systems},
  author={Mozannar, Hussein and Bansal, Gagan and Tan, Cheng and Fourney, Adam and Dibia, Victor and Chen, Jingya and Gerrits, Jack and Payne, Tyler and Maldaner, Matheus Kunzler and Grunde-McLaughlin, Madeleine and others},
  journal={arXiv preprint arXiv:2507.22358},
  year={2025}
}

@article{hao2025uncertainty,
  title={Uncertainty-aware gui agent: Adaptive perception through component recommendation and human-in-the-loop refinement},
  author={Hao, Chao and Wang, Shuai and Zhou, Kaiwen},
  journal={arXiv preprint arXiv:2508.04025},
  year={2025}
}

@article{long2025doubleagents,
  title={DoubleAgents: Interactive Simulations for Alignment in Agentic AI},
  author={Long, Tao and Zhang, Xuanming and Wang, Sitong and Yu, Zhou and Chilton, Lydia B},
  journal={arXiv preprint arXiv:2509.12626},
  year={2025}
}

@article{tsai2026uncertain,
  title={Uncertain Pointer: Situated Feedforward Visualizations for Ambiguity-Aware AR Target Selection},
  author={Tsai, Ching-Yi and Tacconi, Nicole and Wilson, Andrew D and Abtahi, Parastoo},
  journal={arXiv preprint arXiv:2602.13433},
  year={2026}
}

@inproceedings{maquil2023establishing,
  author       = {Val{\'{e}}rie Maquil and
                  Dimitra Anastasiou and
                  Hoorieh Afkari and
                  Adrien Coppens and
                  Johannes Hermen and
                  Lou Schwartz},
  editor       = {Albrecht Schmidt and
                  Kaisa V{\"{a}}{\"{a}}n{\"{a}}nen and
                  Tesh Goyal and
                  Per Ola Kristensson and
                  Anicia Peters},
  title        = {Establishing Awareness through Pointing Gestures during Collaborative
                  Decision-Making in a Wall-Display Environment},
  booktitle    = {Extended Abstracts of the 2023 {CHI} Conference on Human Factors in
                  Computing Systems, {CHI} {EA} 2023, Hamburg, Germany, April 23-28,
                  2023},
  pages        = {104:1--104:7},
  publisher    = {{ACM}},
  year         = {2023}
}

@inproceedings{yen2025code,
  author       = {Ryan Yen and
                  Jian Zhao and
                  Daniel Vogel},
  editor       = {Naomi Yamashita and
                  Vanessa Evers and
                  Koji Yatani and
                  Sharon Xianghua Ding and
                  Bongshin Lee and
                  Marshini Chetty and
                  Phoebe O. Toups Dugas},
  title        = {Code Shaping: Iterative Code Editing with Free-form AI-Interpreted
                  Sketching},
  booktitle    = {Proceedings of the 2025 {CHI} Conference on Human Factors in Computing
                  Systems, {CHI} 2025, YokohamaJapan, 26 April 2025- 1 May 2025},
  pages        = {872:1--872:17},
  publisher    = {{ACM}},
  year         = {2025},
  url          = {https://doi.org/10.1145/3706598.3713822},
  doi          = {10.1145/3706598.3713822}
}

@article{wen2025exploring,
  author       = {Zhen Wen and
                  Luoxuan Weng and
                  Yinghao Tang and
                  Runjin Zhang and
                  Yuxin Liu and
                  Bo Pan and
                  Minfeng Zhu and
                  Wei Chen},
  title        = {Exploring Multimodal Prompt for Visualization Authoring with Large
                  Language Models},
  journal      = {CoRR},
  volume       = {abs/2504.13700},
  year         = {2025},
  url          = {https://doi.org/10.48550/arXiv.2504.13700},
  doi          = {10.48550/ARXIV.2504.13700}
}

@inproceedings{shtedritski2023does,
  author       = {Aleksandar Shtedritski and
                  Christian Rupprecht and
                  Andrea Vedaldi},
  title        = {What does {CLIP} know about a red circle? Visual prompt engineering
                  for VLMs},
  booktitle    = {{IEEE/CVF} International Conference on Computer Vision, {ICCV} 2023,
                  Paris, France, October 1-6, 2023},
  pages        = {11953--11963},
  publisher    = {{IEEE}},
  year         = {2023},
  url          = {https://doi.org/10.1109/ICCV51070.2023.01101},
  doi          = {10.1109/ICCV51070.2023.01101}
}

@inproceedings{cai2024vip,
  author       = {Mu Cai and
                  Haotian Liu and
                  Siva Karthik Mustikovela and
                  Gregory P. Meyer and
                  Yuning Chai and
                  Dennis Park and
                  Yong Jae Lee},
  title        = {ViP-LLaVA: Making Large Multimodal Models Understand Arbitrary Visual
                  Prompts},
  booktitle    = {{IEEE/CVF} Conference on Computer Vision and Pattern Recognition,
                  {CVPR} 2024, Seattle, WA, USA, June 16-22, 2024},
  pages        = {12914--12923},
  publisher    = {{IEEE}},
  year         = {2024},
  url          = {https://doi.org/10.1109/CVPR52733.2024.01227},
  doi          = {10.1109/CVPR52733.2024.01227}
}

@article{cheng2025navi,
  title={Navi-plus: Managing Ambiguous GUI Navigation Tasks with Follow-up Questions},
  author={Cheng, Ziming and Huang, Zhiyuan and Pan, Junting and Hou, Zhaohui and Zhan, Mingjie},
  journal={arXiv preprint arXiv:2503.24180},
  year={2025}
}

@inproceedings{peng2025morae,
  author       = {Yi{-}Hao Peng and
                  Dingzeyu Li and
                  Jeffrey P. Bigham and
                  Amy Pavel},
  editor       = {Andrea Bianchi and
                  Elena L. Glassman and
                  Wendy E. Mackay and
                  Shengdong Zhao and
                  Jeeeun Kim and
                  Ian Oakley},
  title        = {Morae: Proactively Pausing {UI} Agents for User Choices},
  booktitle    = {Proceedings of the 38th Annual {ACM} Symposium on User Interface Software
                  and Technology, {UIST} 2025, Busan, Korea, 28 September 2025 - 1 October
                  2025},
  pages        = {198:1--198:14},
  publisher    = {{ACM}},
  year         = {2025},
  url          = {https://doi.org/10.1145/3746059.3747797},
  doi          = {10.1145/3746059.3747797}
}

@article{yun2025interaction,
  author       = {Hyeonggeun Yun and
                  Jinkyu Jang},
  title        = {Interaction-Driven Browsing: {A} Human-in-the-Loop Conceptual Framework
                  Informed by Human Web Browsing for Browser-Using Agents},
  journal      = {CoRR},
  volume       = {abs/2509.12049},
  year         = {2025},
  url          = {https://doi.org/10.48550/arXiv.2509.12049},
  doi          = {10.48550/ARXIV.2509.12049}
}

@article{tang2026human,
  title={Human Tool: An MCP-Style Framework for Human-Agent Collaboration},
  author={Tang, Yuanrong and Peng, Huiling and Zhao, Bingxi and Ding, Hengyang and Song, Hanchao and Wang, Tianhong and Zhong, Chen and Gong, Jiangtao},
  journal={arXiv preprint arXiv:2602.12953},
  year={2026}
}

@article{wu2025gui,
  title={Gui-reflection: Empowering multimodal gui models with self-reflection behavior},
  author={Wu, Penghao and Ma, Shengnan and Wang, Bo and Yu, Jiaheng and Lu, Lewei and Liu, Ziwei},
  journal={Advances in Neural Information Processing Systems},
  volume={38},
  pages={101861--101896},
  year={2026}
}

@inproceedings{sun2025genesis,
  author       = {Qiushi Sun and
                  Kanzhi Cheng and
                  Zichen Ding and
                  Chuanyang Jin and
                  Yian Wang and
                  Fangzhi Xu and
                  Zhenyu Wu and
                  Chengyou Jia and
                  Liheng Chen and
                  Zhoumianze Liu and
                  Ben Kao and
                  Guohao Li and
                  Junxian He and
                  Yu Qiao and
                  Zhiyong Wu},
  editor       = {Wanxiang Che and
                  Joyce Nabende and
                  Ekaterina Shutova and
                  Mohammad Taher Pilehvar},
  title        = {OS-Genesis: Automating {GUI} Agent Trajectory Construction via Reverse
                  Task Synthesis},
  booktitle    = {Proceedings of the 63rd Annual Meeting of the Association for Computational
                  Linguistics (Volume 1: Long Papers), {ACL} 2025, Vienna, Austria,
                  July 27 - August 1, 2025},
  pages        = {5555--5579},
  publisher    = {Association for Computational Linguistics},
  year         = {2025}
}

@article{guo2025auto,
  author       = {Xiangwu Guo and
                  Difei Gao and
                  Mike Zheng Shou},
  title        = {AUTO-Explorer: Automated Data Collection for {GUI} Agent},
  journal      = {CoRR},
  volume       = {abs/2511.06417},
  year         = {2025},
  url          = {https://doi.org/10.48550/arXiv.2511.06417},
  doi          = {10.48550/ARXIV.2511.06417},
  eprinttype   = {arXiv},
  eprint       = {2511.06417},
  bibsource    = {dblp computer science bibliography, https://dblp.org}
}

@article{kang2026learning,
  title={Learning with Challenges: Adaptive Difficulty-Aware Data Generation for Mobile GUI Agent Training},
  author={Kang, Linjia and Wang, Zhimin and Zhang, Yongkang and Wu, Duo and Wang, Jinghe and Ma, Ming and Yan, Haopeng and Wang, Zhi},
  journal={arXiv preprint arXiv:2601.22781},
  year={2026}
}

@article{wang2004image,
  author       = {Zhou Wang and
                  Alan C. Bovik and
                  Hamid R. Sheikh and
                  Eero P. Simoncelli},
  title        = {Image quality assessment: from error visibility to structural similarity},
  journal      = {{IEEE} Trans. Image Process.},
  volume       = {13},
  number       = {4},
  pages        = {600--612},
  year         = {2004}
}

@inproceedings{
madden2024robustness,
title={Robustness of Practical Perceptual Hashing Algorithms to Hash-Evasion and Hash-Inversion Attacks},
author={Jordan Madden and Moxanki Bhavsar and Lhamo Dorje and Xiaohua Li},
booktitle={The Third Workshop on New Frontiers in Adversarial Machine Learning},
year={2024},
url={https://openreview.net/forum?id=hraOxsleRl}
}

@article{jia2025your,
  author       = {Allison Sihan Jia and
                  Daniel Huang and
                  Nikhil Vytla and
                  Nirvika Choudhury and
                  Shayak Sen and
                  John C. Mitchell and
                  Anupam Datta},
  title        = {What Is Your Agent's GPA? {A} Framework for Evaluating Agent
                  Goal-Plan-Action Alignment},
  journal      = {CoRR},
  volume       = {abs/2510.08847},
  year         = {2025}
}

@inproceedings{grigorev2025verifyllm,
  author       = {Danil S. Grigorev and
                  Alexey K. Kovalev and
                  Aleksandr I. Panov},
  title        = {VerifyLLM: LLM-Based Pre-Execution Task Plan Verification for Robots},
  booktitle    = {{IEEE/RSJ} International Conference on Intelligent Robots and Systems,
                  {IROS} 2025, Hangzhou, China, October 19-25, 2025},
  pages        = {18489--18496},
  publisher    = {{IEEE}},
  year         = {2025}
}

@inproceedings{balepur2025good,
  author       = {Nishant Balepur and
                  Matthew Shu and
                  Yoo Yeon Sung and
                  Seraphina Goldfarb{-}Tarrant and
                  Shi Feng and
                  Fumeng Yang and
                  Rachel Rudinger and
                  Jordan Lee Boyd{-}Graber},
  editor       = {Christos Christodoulopoulos and
                  Tanmoy Chakraborty and
                  Carolyn Rose and
                  Violet Peng},
  title        = {A Good Plan is Hard to Find: Aligning Models with Preferences is Misaligned
                  with What Helps Users},
  booktitle    = {Proceedings of the 2025 Conference on Empirical Methods in Natural
                  Language Processing, {EMNLP} 2025, Suzhou, China, November 4-9, 2025},
  pages        = {11568--11595},
  publisher    = {Association for Computational Linguistics},
  year         = {2025}
}

@inproceedings{wei2025plangenllms,
  author       = {Hui Wei and
                  Zihao Zhang and
                  Shenghua He and
                  Tian Xia and
                  Shijia Pan and
                  Fei Liu},
  editor       = {Wanxiang Che and
                  Joyce Nabende and
                  Ekaterina Shutova and
                  Mohammad Taher Pilehvar},
  title        = {PlanGenLLMs: {A} Modern Survey of {LLM} Planning Capabilities},
  booktitle    = {Proceedings of the 63rd Annual Meeting of the Association for Computational
                  Linguistics (Volume 1: Long Papers), {ACL} 2025, Vienna, Austria,
                  July 27 - August 1, 2025},
  pages        = {19497--19521},
  publisher    = {Association for Computational Linguistics},
  year         = {2025}
}
